\documentclass[opew,nonblindrev]{informs3}
\usepackage[T1]{fontenc}
\usepackage[utf8]{inputenc}

\OneAndAHalfSpacedXI

\usepackage{endnotes}
\let\footnote=\endnote
\let\enotesize=\normalsize
\def\notesname{Endnotes}%
\def\makeenmark{\hbox to1.275em{\theenmark.\enskip\hss}}
\def\enoteformat{\rightskip0pt\leftskip0pt\parindent=1.275em \leavevmode\llap{\makeenmark}}

\usepackage{amsmath,amsfonts,amssymb,bm,dsfont,lmodern}
\usepackage{enumitem}
\usepackage{multirow}
\usepackage{float}
\usepackage{epstopdf}

\usepackage{array}
\usepackage{graphicx}
\usepackage{caption}
\usepackage{subcaption}
\usepackage{algorithm}
\usepackage{algpseudocode}
\algdef{SE}[SUBALG]{Indent}{EndIndent}{}{\algorithmicend\ }%
\algtext*{Indent}
\algtext*{EndIndent}

\usepackage{hyperref}
\newcommand{\thmref}[2]{\hyperref[#1]{#2 \ref*{#1}}}
\usepackage[dvipsnames]{xcolor}
\newcommand\myshade{100}
\colorlet{mylinkcolor}{MidnightBlue}
\hypersetup{
linkcolor  = mylinkcolor!\myshade!black,
citecolor  = mylinkcolor!\myshade!black,
urlcolor   = mylinkcolor!\myshade!black,
colorlinks = true,
}

\usepackage{titlesec}
\usepackage[titletoc,toc,title]{appendix}

\renewcommand{\arraystretch}{1.5}

\usepackage{tabularx}
\usepackage{pict2e}
\newcolumntype{L}[1]{>{\raggedright\arraybackslash}p{#1}}
\newcolumntype{C}[1]{>{\centering\arraybackslash}p{#1}}
\newcolumntype{R}[1]{>{\raggedleft\arraybackslash}p{#1}}

\usepackage{thmtools}
\usepackage{thm-restate}

\usepackage{cleveref}
\usepackage{threeparttable}

\usepackage{natbib}
\bibpunct[, ]{(}{)}{,}{a}{}{,}%
\def\bibfont{\small}%

\TheoremsNumberedThrough 
\EquationsNumberedThrough 
\ECRepeatTheorems

\usepackage{color-edits}
\addauthor{rv}{blue}
\addauthor{zc}{violet}
\addauthor{jf}{ForestGreen}

\usepackage{booktabs}
\usepackage{graphicx}     
\usepackage{array}        
\usepackage{makecell}     
\usepackage{booktabs}     
\usepackage[most]{tcolorbox} 
\usepackage{xcolor}        
\usepackage{amsmath}       
\usepackage{caption}       
\usepackage{hyperref}      
\usepackage{cleveref} 
\usepackage{longtable}
\usepackage{tabularx}

\begin{document}
\sloppy

\TITLE{Large Language Newsvendor: \\ Decision Biases and Cognitive Mechanisms}
\RUNTITLE{Large Language Newsvendor: Decision Biases and Cognitive Mechanisms}
\RUNAUTHOR{Liu, Chen, and Zhong}

\ARTICLEAUTHORS{%
\AUTHOR{Jifei Liu$^1$, Zhi Chen$^2$, and Yuanguang Zhong$^1$}
\AFF{$^1$School of Business Administration, South China University of Technology\\
202510189129@mail.scut.edu.cn; bmygzhong@scut.edu.cn}
\AFF{$^2$Department of Decisions, Operations and Technology, The Chinese University of Hong Kong\\
zhi.chen@cuhk.edu.hk}
}

\ABSTRACT{{\textit{Problem definition:} Although large language models (LLMs) are increasingly integrated into business decision making, their potential to replicate and even amplify human cognitive biases cautions a significant, yet not well-understood, risk. This is particularly critical in high-stakes operational contexts like supply chain management. To address this, we investigate the decision-making patterns of leading LLMs using the canonical newsvendor problem in a dynamic setting, aiming to identify the nature and origins of their cognitive biases. \textit{Methodology/results:} Through dynamic, multi-round experiments with GPT-4, GPT-4o, and LLaMA-8B, we tested for five established decision biases. We found that LLMs consistently replicated the classic ``Too Low/Too High'' ordering bias and significantly amplified other tendencies like demand-chasing behavior compared to human benchmarks. Our analysis uncovered a ``paradox of intelligence'': the more sophisticated GPT-4 demonstrated the greatest irrationality through overthinking, while the efficiency-optimized GPT-4o performed near-optimally. Because these biases persist even when optimal formulas are provided, we conclude they stem from architectural constraints rather than knowledge gaps. \textit{Managerial implications:} First, managers should select models based on the specific task, as our results show that efficiency-optimized models can outperform more complex ones on certain optimization problems. Second, the significant amplification of bias by LLMs highlights the urgent need for robust human-in-the-loop oversight in high-stakes decisions to prevent costly errors. Third, our findings suggest that designing structured, rule-based prompts is a practical and effective strategy for managers to constrain models' heuristic tendencies and improve the reliability of AI-assisted decisions.}}


\HISTORY{\today}
\maketitle
\section{Introduction}\label{sec:introduction}

Artificial intelligence (AI) has transformed economic decision making (\citealt{agrawal2022prediction}), with significant advancements in supply chain management. The global market for AI in supply chains was valued at \$9.15 billion in 2024 and is projected to reach \$40.53 billion by 2030, reflecting a compound annual growth rate of 28.2\%.\footnotemark\footnotetext{MarketsandMarkets, ``AI in Supply Chain Market worth \$40.53 billion by 2030'', 2024. \url{https://www.marketsandmarkets.com/Market-Reports/ai-in-supply-chain-market-114588383.html}.} This growth underscores AI's pivotal role in optimizing inventory, forecasting demand, and coordinating logistics for major corporations. Quite notably, over 92\% of Fortune 500 companies have adopted AI tools,\footnotemark\footnotetext{Tech Business News, ``92\% of Fortune 500 Companies Use OpenAI Products'', 2025. \url{https://www.techbusinessnews.com.au/news/92-of-fortune-500-companies-use-openai-products/}.} and 67\% of businesses worldwide now leverage generative AI systems,\footnotemark\footnotetext{Market.us, ``Large Language Model (LLM) Market'', 2025. \url{https://market.us/report/large-language-model-llm-market/}.} particularly those powered by Large Language Models (LLMs), to enhance operational efficiency.

The economic benefits of AI in supply chains are substantial: AI applications can reduce forecasting errors by up to 50\%, decrease lost sales by 65\%, and lower end-to-end supply chain costs by an average of 25\%.\footnotemark\footnotetext{McKinsey \& Company, ``AI-driven Operations Forecasting in Data-light Environments'', 2022. \url{https://www.mckinsey.com/capabilities/operations/our-insights}.} Leading firms illustrate these advantages: Amazon employs AI to analyze customer feedback for constant improvement, Zara uses AI to optimize inventory and identify market trends, and Siemens automates logistics processes, from demand forecasting to real-time inventory tracking. Studies demonstrate that companies achieve an average return of \$3.50 for every \$1 invested in AI, with 49\% of organizations expecting to realize ROI within one to three years of implementation.\footnotemark\footnotetext{Pecan AI, ``How to Measure (and Increase) the ROI of AI Initiatives'', 2024. \url{https://www.pecan.ai/blog/how-to-measure-increase-roi-of-ai/}.}

However, the integration of LLMs into decision-making systems raises critical concerns. Trained on vast datasets of human-generated text, LLMs may inherit or amplify cognitive biases observed in human decision making, rather than achieving fully rational outcomes. Decades of behavioral economics research have identified systematic deviations from rationality, including anchoring effects (\citealt{kahneman1982judgment}), overconfidence (\citealt{ren2013overconfidence}), and loss aversion (\citealt{kahneman2013prospect}). These biases significantly influence professional decisions in fields such as management, finance, medicine, and law. When embedded in AI systems managing high-stakes decisions, such biases could lead to substantial economic consequences. These concerns are well-supported by emerging evidence. Recent studies indicate that LLMs exhibit decision-making patterns analogous to human cognitive biases: GPT-3 displays loss aversion, probability weighting, and overconfidence (\citealt{hagendorff2023human, binz2023using, talboy2023challenging}), while also showing social preferences in economic contexts (\citealt{filippas2024large, brand2023using}). With ChatGPT attracting over 3 billion monthly visits by late 2024---compared to 300 million for Google's Gemini---understanding these biases is critical for business applications.

Despite these findings, current research has notable limitations with practical implications. Most studies rely on single-round experiments, failing to capture how biases evolve under dynamic feedback---a key characteristic of real-world operations. With 62\% of Chief Supply Chain Officers expecting generative AI to accelerate discovery and innovation,\footnotemark\footnotetext{IBM Institute for Business Value, ``The Intuitive, AI-powered Supply Chain'', 2024. \url{https://www.ibm.com/thought-leadership/institute-business-value/en-us/report/supply-chain-ai}.} there is a lack of systematic comparisons across LLM architectures in canonical problems with established human behavioral benchmarks. Although \cite{chen2025manager} demonstrated consistent bias patterns in GPT models within specific operational contexts, however, the specific ways these biases manifest across different LLM architectures and evolve under dynamic feedback remain underexplored.

To address these gaps, this study investigates the newsvendor problem---a foundational model in inventory management with analytical optimum, extensive human behavioral data spanning seven decades, and direct relevance to AI-driven operations shaping global commerce. We systematically evaluate three leading LLMs---GPT-4, GPT-4o, and LLaMA-8B---across varied demand distributions in dynamic, multi-round experiments that incorporate feedback and learning. Our analysis focuses on five key decision biases: systematic ordering bias, presentation-order effect, bias persistence in the risk-neutral setting, demand-chasing behavior, and constraints on learning from feedback.

Our findings reveal that LLMs exhibit distinct decision-making patterns shaped by their underlying architectures. All models replicated the classic ``Too Low/Too High'' ordering bias observed in human behavior, with GPT-4 deviating up to 70\% more than human benchmarks. These deviations persisted even in risk-neutral environments, indicating that they stem from architectural constraints rather than risk aversion. Surprisingly, we uncovered a counterintuitive trend: the most computationally advanced model demonstrated the greatest deviations from optimality due to overcomplex reasoning, while the efficiency-optimized GPT-4o achieved near-optimal performance. This challenges the prevailing assumption that increased model sophistication leads to more rational decision making.

This study offers critical insights for the \$40.53 billion AI supply chain market projected for 2030. By treating LLMs as experimental subjects for behavioral analysis, our study questions the notion that increased model complexity inherently improves decision quality and advances the understanding of artificial cognition. By identifying specific mechanisms through which cognitive biases manifest in LLM decision making, our findings provide actionable guidance for designing robust AI-assisted inventory systems and implementing effective human oversight, particularly in contexts where bias amplification could lead to significant economic losses.

The remainder of this paper is organized as follows. Section~\ref{sec: results review} reviews the literature on cognitive biases in human and LLM decision making. Section~\ref{sec: hypotheses} outlines our hypotheses. Section~\ref{sec:experiment} details the experimental design and methodology, and Section~\ref{sec: results} presents and analyzes the empirical findings. Finally, Section~\ref{sec:discussion} discusses and concludes this study.

\section{Literature Review}\label{sec: results review}

To understand how LLMs make economic decisions, it is essential to examine the landscape of human cognitive biases and explore their potential emergence in artificial systems. We review the extensive literature on human judgment biases in economic and operational contexts and the emerging research on analogous patterns by LLMs, highlighting the critical research gaps that our study addresses.

\subsection{Cognitive Biases in Human Decision Making}

Human decision making exhibits consistent and predictable deviations from rational models, a discovery that has transformed economic theory over recent decades (\citealp{kahneman1982judgment}). Foundational frameworks, such as the Allais paradox (\citealp{allais1953comportement}) and prospect theory (\citealp{kahneman2013prospect}), demonstrate that cognitive biases are integral to human information processing rather than isolated anomalies (\citealp{simon1955behavioral, gigerenzer2011heuristic}). Within operational contexts, the newsvendor problem has become a cornerstone for examining such biases (\citealp{schweitzer2000decision}). One of the most well-documented phenomena is the ``Too Low/Too High'' effect, where decision makers tend to underorder in high-margin scenarios and overorder in low-margin scenarios. First identified by \cite{schweitzer2000decision} and subsequently replicated in numerous studies (see, \textit{e.g.}, \citealp{bolton2008learning, bostian2008newsvendor}), this pattern has been attributed to cognitive overconfidence (\citealp{ren2013overconfidence}) and mean-anchoring, in which individuals anchor on expected demand but make insufficient adjustments toward the optimal order quantity (\citealp{chen2013effect}).

Importantly, the persistence of these biases is not limited to situations involving risk of loss. Even in risk-neutral environments where all outcomes are profitable, cognitive biases remain observable (\citealp{bostian2008newsvendor}). This contradicts traditional economic assumptions that such deviations arise primarily from risk aversion and instead suggests that they are rooted in more fundamental cognitive processes, such as anchoring on expected demand (\citealp{chen2013effect}). Furthermore, the sequence in which information is presented can substantially shape decisions. For example, initial exposure to a low-margin scenario fosters conservative ordering strategies that often persist despite subsequent changes in the decision environment (\citealp{bolton2008learning}). This presentation-order effect aligns with broader behavioral phenomena, including anchoring (\citealp{kahneman1982judgment}), path dependence (\citealp{schilling2001use}), and selective information processing (\citealp{hogarth1992order}).

In multi-period settings, decision makers frequently display demand-chasing behavior, overreacting to recent demand realizations while neglecting longer-term demand patterns (\citealp{schweitzer2000decision, niranjan2023unpacking}). Such behavior persists even among experienced practitioners and reflects reliance on heuristic shortcuts, such as the law of small numbers (\citealp{camerer1999experience}) and other cognitive heuristics (\citealp{gigerenzer2011heuristic}). While learning from feedback can lead to some performance improvements (\citealp{bolton2008learning}), decision quality often plateaus at suboptimal levels. These limitations in learning are compounded by the resilience of biases, which frequently resist correction even when feedback is comprehensive (\citealp{bostian2008newsvendor}). Additional barriers, such as hindsight bias (\citealp{kahneman1982judgment}) and outcome bias (\citealp{baron1988outcome}), further restrict the effectiveness of experiential learning.

\subsection{Cognitive Biases in Large Language Models}

LLMs have demonstrated exceptional performance in complex reasoning tasks, often rivaling or surpassing human capabilities (\citealp{wei2022chain, bubeck2023sparks}). However, the fact that they are trained from extensive human-generated text datasets raises a critical question: do they adopt cognitive biases inherent in their training data alongside factual knowledge?

Emerging research suggests that LLMs exhibit decision-making patterns analogous to human cognitive biases. Studies have shown that GPT-3 displays loss aversion and probability weighting (\citealp{hagendorff2023human, binz2023using}), while other analyses indicate that LLMs may exhibit and potentially intensify biases such as overconfidence (\citealp{talboy2023challenging}). A comprehensive study by \cite{malberg2024comprehensive} evaluated 20 mainstream LLMs across 30 measures of bias, finding that every model exhibited significant bias in at least one of the tests. These findings highlight a pervasive challenge in LLM decision making. Encouragingly, targeted interventions to enhance rational decision making have shown promise, with \cite{sumita2025cognitive} demonstrating that specific debiasing techniques improve the rationality of responses in GPT-3.5 and GPT-4.

In economic contexts, LLMs show cognitive tendencies mirroring human behavior, such as social preferences in economic games (\citealp{filippas2024large}), simulating experimental populations (\citealp{aher2023using}), and consistent deviations under uncertainty (\citealp{brand2023using}). Most relevant to this study, \cite{chen2025manager} conducted a comprehensive analysis of GPT models within operations management contexts. Their study tested GPT-3.5 and GPT-4 against 18 human biases using single-shot decision vignettes. A primary finding of their work is the ``dual-edged progression'' of LLM capabilities, wherein GPT-4 demonstrates higher accuracy on tasks with objective solutions, while simultaneously exhibiting stronger human-like biases in preference-based tasks. This supports the theory that an LLM’s sampling behavior is influenced by a combination of descriptive norms—what is statistically likely—and prescriptive norms—what is considered ideal. The systematic deviation from the average toward an ideal, as demonstrated by \cite{sivaprasad-etal-2025-theory}, provides a potent explanatory framework for the ordering biases observed in operational tasks like the newsvendor problem. However, the research methodologies in these studies, including the work of \cite{chen2025manager}, often rely on static, one-off experiments. This approach, while broad in scope, does not capture the dynamic nature of decision-making processes that unfold over time.

\subsection{Research Gaps}

Despite substantial progress, research on LLM decision making in operational contexts remains unexplored in several important ways. First of all, many existing studies rely on static, single-round experiments that capture only momentary instances of bias, whereas real-world operations are characterized by dynamic feedback loops, iterative learning, and evolving conditions. Although a recent work by \cite{chen2025manager} has begun to examine LLM biases in operational settings, the focus remains largely on single-shot decision scenarios. While emerging studies on LLM-based agents highlight their potential for dynamic interaction (\citealp{wang2024survey, park2023generative}), little is known about how such biases evolve over time or respond to feedback---an essential consideration given the iterative nature of economic and operational decision making.

Another limitation lies in the absence of systematic comparisons across different LLM architectures and versions. Research in strategic decision making demonstrates substantial performance differences across model architectures and training paradigms (\citealp{csaszar2024artificial}), and similar variation may influence the expression of biases in operational contexts (\citealp{mirza2025gender}). The rapid emergence of mixture-of-experts designs and other architectural innovations further underscores the importance of understanding how these differences affect decision-making patterns, a prerequisite for developing robust AI systems suitable for deployment in real-world operations.

A further gap concerns the limited use of canonical operational models with analytical optima and extensive human behavioral benchmarks. Although standardized evaluation frameworks are gaining traction in domains such as clinical decision making (\citealp{park2024assessing}) and strategic contexts (\citealp{csaszar2024artificial}), operational settings still lack comparable approaches. Models like the newsvendor problem offer a unique combination of mathematical rigor (\citealp{arrow1951optimal}) and rich human behavioral datasets (\citealp{schweitzer2000decision}), yet remain underutilized for LLM assessment. Without such benchmarks, it is difficult to determine whether LLM biases parallel, diverge from, or exceed human biases, or to evaluate how varying uncertainty structures influence AI decision behavior (\citealp{jia2024decision, da2025understanding}).

To address these gaps, we conduct an in-depth experiment on LLM decision making in the newsvendor problem, the canonical model in supply chain management. Our dynamic, multi-round experimental design incorporates feedback and learning across multiple LLM architectures (GPT-4, GPT-4o, and LLaMA-8B) and varied demand distributions. This approach enables tracking the evolution of bias across sequential decisions, facilitates systematic comparisons across models, and grounds the results in several decades of human behavioral research. Focusing on five documented decision biases---systematic ordering bias, presentation-order effect, bias persistence in the risk-neutral setting, demand-chasing behavior, and constraints on learning from feedback---our work connects advanced LLM research with classical behavioral operations management to reveal both human-like and distinct AI-driven cognitive patterns in operational decision making.

\section{Hypotheses}\label{sec: hypotheses}

Building on established research on human cognitive biases and emerging evidence on LLM decision making, we develop five hypotheses to investigate how LLMs behave in the newsvendor problem. These hypotheses span well-documented behavioral deviations, contextual influences, and dynamic learning patterns, aiming to bridge insights from behavioral operations management with current advances in AI research.

\subsection{Systematic Ordering Bias}

Among the most robust findings in behavioral operations is the ``Too Low/Too High'' ordering pattern, in which decision makers systematically underorder in high-margin scenarios and overorder in low-margin scenarios (\citealp{schweitzer2000decision}). This deviation has been observed across laboratory and field studies (\citealp{bolton2008learning, bostian2008newsvendor, ho2010reference}) and is often linked to cognitive overconfidence (\citealp{ren2013overconfidence}) or mean-anchoring, anchoring on expected demand with insufficient adjustment (\citealp{chen2013effect}). This bias persists even among professionals, indicating deep-seated cognitive tendencies. Since LLMs are trained on large-scale human-generated corpora, likely encoding such patterns, we expect them to reproduce this fundamental deviation.
\begin{hypothesis}\label{hypo:1}
\textit{Without access to explicit optimal-ordering formulas, LLMs underestimate optimal order quantities in high-margin scenarios and overestimate them in low-margin scenarios, replicating the ``Too Low/Too High'' deviation observed in human decision makers.}
\end{hypothesis}

\subsection{Bias Persistence in Risk-Neutral Setting}
Classical economic models explain ordering deviations as resulting from risk aversion, driven by the perceived cost of excess inventory. However, behavioral studies show that such deviations persist even in risk-neutral environments where all possible outcomes yield positive profits (\citealp{nagarajan2014prospect, narayanan2015decision}). This suggests underlying cognitive processes, such as anchoring and heuristic-driven reasoning (\citealp{kremer2010random, li2017overconfident}), rather than purely preference-based explanations. For LLMs, this distinction is important: if deviations stem from preferences over risk, they may diminish in risk-neutral contexts; if they originate from more fundamental information-processing tendencies, they should remain.
\begin{hypothesis}\label{hypo:2}
\textit{In risk-neutral scenarios where all possible order quantities yield positive profits, LLMs continue to exhibit ordering deviations, indicating that these patterns arise from core information-processing heuristics rather than risk aversion.}
\end{hypothesis}

\subsection{Presentation-Order Effect}
Decision outcomes are shaped not only by problem parameters but also by the sequence in which information is presented. In the newsvendor context, initial exposure to low-margin scenario leads to conservative ordering behavior that often persists even after margins increase (\citealp{bolton2008learning}). Such path dependence reflects broader cognitive mechanisms, including anchoring effects (\citealp{kahneman1982judgment}), selective processing (\citealp{chapman1999anchoring}), and framing effects (\citealp{gavirneni2010anatomy}). LLMs process information sequentially through attention mechanisms (\citealp{vaswani2017attention}), which may make them susceptible to similar contextual anchoring effects.
\begin{hypothesis}\label{hypo:3}
\textit{LLMs’ ordering behavior is influenced by the sequence of margin scenarios, exhibiting a presentation-order effect analogous to that observed in human decision making.}
\end{hypothesis}

\subsection{Demand-Chasing Behavior}
In multi-period newsvendor tasks, human decision makers frequently overreact to recent demand realizations, a tendency known as demand-chasing behavior (\citealp{schweitzer2000decision}). They overweight recent observations rather than relying on long-term demand distributions (\citealp{niranjan2023unpacking}), a bias that persists even among experienced professionals (\citealp{ren2017overconfident}). This behavior is consistent with heuristics such as the law of small numbers (\citealp{kahneman1971belief}), experience-weighted attraction learning (\citealp{camerer1999experience}), and misperceptions of randomness (\citealp{burns2004randomness}). Given that LLMs operate through sequential token prediction and are trained on human text, they may similarly overemphasize recent information in updating decisions.
\begin{hypothesis}\label{hypo:4}
\textit{In multi-round newsvendor tasks, LLMs adjust order quantities disproportionately based on recent demand realizations, exhibiting overreaction to recent signals comparable to or exceeding that of human decision makers.}
\end{hypothesis}

\subsection{Constraints on Learning from Feedback}
While feedback can improve human decision making in the newsvendor problem, performance gains typically plateau at suboptimal levels (\citealp{bolton2008learning}). Deviations persist despite rich and immediate feedback (\citealp{bostian2008newsvendor}), even among seasoned practitioners (\citealp{feng2017modeling}). This resilience of bias is attributed to cognitive barriers such as hindsight bias (\citealp{kahneman1982judgment}), outcome bias (\citealp{einhorn1978confidence}), and limitations in processing probabilistic information, with feedback format and frequency sometimes producing counterintuitive effects (\citealp{lurie2009timely}). Given the architectural differences among LLMs, their learning from feedback is likely variable but, as in humans, may not be sufficient to fully overcome entrenched deviations.
\begin{hypothesis}\label{hypo:5}
\textit{When provided with decision feedback, LLMs exhibit behavioral adjustments, yet these adjustments are insufficient to eliminate persistent ordering deviations, reflecting learning constraints akin to those found in human decision makers.}
\end{hypothesis}

\section{Details of Our Experiment}\label{sec:experiment}

Testing these hypotheses requires a rigorously designed experimental framework to capture the nuanced decision-making processes of LLMs under uncertainty. This section outlines our approach, detailing the selected models, experimental conditions, and data collection methods. Our aim is to establish a controlled yet realistic environment that enables systematic testing and analysis of LLM decision making while ensuring the rigor necessary for robust scientific insights.

\subsection{Experimental Setup}

We selected three LLMs with distinct capabilities: GPT-4, GPT-4o, and LLaMA-8B. Each model offers a distinct perspective on how architectural variations impact decision making. GPT-4, known for its advanced reasoning capabilities, excels in deep contextual understanding and complex logical processing. GPT-4o, optimized for computational efficiency, maintains robust performance, offering insights into streamlined architectures. LLaMA-8B, with its smaller scale, enables analysis of how computational constraints affect ordering tendencies. We accessed GPT-4 and GPT-4o via the OpenAI API\footnotemark\footnotetext{OpenAI: \url{https://openai.com/api/}} to ensure consistency and reproducibility, while deploying LLaMA-8B locally. To conduct experiments systematically, we developed an automated Python-based platform\footnotemark\footnotetext{The code for our automated experimental platform is publicly available on GitHub: \url{ https://github.com/kingsfei/llm-newsvendor-experiment.git}} that managed model interactions, recorded decisions and rationales, and enforced strict experimental control. A key design is to set the temperature parameter to 1.0 for all models, introducing controlled variability to reflect human-like decision variability while ensuring consistency for robust analysis.

In the newsvendor problem, the decision maker selects an order quantity $q$ before the selling season, given a unit cost $c$ and a unit price $p$. Denote the realized demand by $d$. If $q \geq d$, excess inventory is salvaged with no value; if $q < d$, sales opportunities are lost. The realized profit is:
\[
\pi(q, d) = p  \cdot \min\{q, d\} - c \cdot q.
\]
The optimal order quantity $q^\star$, maximizing the expected profit $\mathbb{E}[\pi(q, \tilde{d})]$ with respect to the distribution $F$ of random demand $\tilde{d}$, satisfies:
\[
\mathbb{P}[\tilde{d} \leq q^\star] = \frac{p - c}{p} := \eta ~~~\Longleftrightarrow~~~ q^\star = F^{-1}(\eta).
\]
This ratio, known as the critical fractile, determines the optimal balance between overordering and underordering, distinguishing high-margin (\textit{i.e.}, $\eta \geq 0.5$) and low-margin (\textit{i.e.}, $\eta < 0.5$) scenarios.

We extend the classical newsvendor framework introduced by \cite{schweitzer2000decision} by incorporating varied demand distributions as a key experimental variable, enabling a comprehensive analysis of LLM decision making under uncertainty.

\begin{itemize}
    \item \textbf{Uniform Distribution $\mathbb{U}[a,b]$}. $\;$ This distribution has a constant probability density and has been employed by \cite{schweitzer2000decision} in their well-known experiment with human decision makers. It has a constant probability density over the interval $[a,b]$.
    
    \item \textbf{Truncated Normal Distribution $\mathbb{N}(\mu_N, \sigma_N^2)$}. $\;$ To better reflect the tendency of real-world demand concentration, we truncate a normal distribution with a mean $\mu_N = (a+b)/2$ and a standard deviation $\sigma_N = (b-a)/6$. This ensures that approximately $99.7\%$ of the untruncated distribution falls within the experimental demand range $[a,b]$.
    
    \item \textbf{Lognormal Distribution}. $\;$ It exhibits positive skewness and is frequently observed in economic contexts, such as bestseller sales. Parameters are fitted to ensure that most probability mass falls within the range $[a,b]$, with the mean reasonably close to $(a+b)/2$.
\end{itemize}

Our experimental design, outlined in Table~\ref{tab:conditions}, progresses with three experiments as follows.

\begin{itemize}
    \item \textbf{Experiment~1 (Baseline)}. $\;$ We test LLMs without theoretical guidance, consistent with the original human experiments (\citealt{schweitzer2000decision}), to establish initial ordering deviations of LLMs driven by the models’ intrinsic training.
    
    \item \textbf{Experiment~2 (with Optimal Formula)}. $\;$ We provide the mathematically optimal formula to evaluate the models’ ability to apply explicit guidance, testing the distinction between theoretical knowledge and practical implementation.
    
    \item \textbf{Experiment~3 (Risk-Neutral Environment)}. $\;$ To isolate biases from risk preferences, we maintain the cost structure but shift the demand range to $[901, 1200]$, ensuring profitability. Persistent deviations in this setting would indicate cognitive rather than risk-driven influences.
\end{itemize}

\begin{table}[tb]
\centering
\footnotesize
\caption{Experimental Configurations}
\label{tab:conditions}
\renewcommand{\arraystretch}{1.2}
\begin{tabular}{
>{\raggedright\arraybackslash}p{4.5cm}
>{\centering\arraybackslash}p{2.2cm}
>{\centering\arraybackslash}p{2.2cm}
>{\centering\arraybackslash}p{1.5cm}
}
\toprule
\makecell[c]{Experimental \\ Condition} & 
\makecell[c]{Cost \\ Structure} & 
\makecell[c]{Demand \\ Range} & 
\makecell[c]{Prompt \\ Formula} \\
\midrule
Experiment~1: High Margin & $c = 3$ & [1, 300] & No \\
Experiment~1: Low Margin & $c = 9$ & [1, 300] & No \\
\cmidrule(lr){1-4}
Experiment~2: High Margin & $c = 3$ & [1, 300] & Yes \\
Experiment~2: Low Margin & $c = 9$ & [1, 300] & Yes \\
\cmidrule(lr){1-4}
\makecell[l]{Experiment~3: High Margin \\ \footnotesize \hspace{2em} (High Demand)} & $c = 3$ & [901, 1200] & Yes \\
\makecell[l]{Experiment~3: Low Margin \\ \footnotesize \hspace{2em} (High Demand)} & $c = 9$ & [901, 1200] & Yes \\
\bottomrule
\multicolumn{4}{p{0.7\textwidth}}{\footnotesize{
\scriptsize{\textit{Notes.} We set $p = 12$ for all experimental conditions.}
}}
\end{tabular}
\end{table}

\subsection{Experimental Procedure and Data Collection}

Unlike most studies that rely on single-shot decisions, we implemented a dynamic 15-round setup to examine the evolution of LLM decision making in the newsvendor problem. This design captures iterative learning, adaptation, and feedback, which are critical in real-world operational contexts but have not been adequately addressed by static experiments.

Each round followed a standardized three-stage procedure: (\textit{i}) giving context and instructions, (\textit{ii}) prompting for an ordering decision with accompanying reasoning, and (\textit{iii}) providing feedback on the previous round’s outcome. This structure parallels the learning and adaptation processes observed in human decision makers in operational settings. 

The following illustrates the Round~1 prompt.

\begin{tcolorbox}[colback=orange!5, colframe=orange!80, title=Prompts (Round~1)]
You are participating in a newsvendor decision experiment. Your task is to decide how many units of ``wodgets'' to order. \\
Key information: the price is 12 francs per unit, the cost is 3 francs per unit, and leftover units have zero salvage value. \\
Demand is uniformly distributed between 1 and 300. \\
You will receive feedback on actual demand and realized profit after each round. Your goal is to maximize the cumulative profit. \\
Please weigh the trade-off between overordering (leftover) and underordering (shortage), make a decision, and explain your reasoning.
\end{tcolorbox}

From Round~2 onward, detailed feedback was provided.

\begin{tcolorbox}[colback=gray!5!white, colframe=gray!50!black, title=Feedback (Round~2 onward)]
Results of the previous round. \\
- You ordered 185 units. \\
- Actual demand was 210 units. \\
- Profit realized: 1665 francs. \\
- Cumulative profit: 1665 francs. \\
Please make your new ordering decision and explain your reasoning.
\end{tcolorbox}

For Experiment~2, the prompt includes the optimal formula to assess the models’ ability to apply theoretical guidance.

\begin{tcolorbox}[colback=blue!5!white, colframe=blue!50!black, title=System Prompt (with Optimal Formula)]
The optimal order quantity $q^\star$ for the newsvendor decision satisfies: \\[3pt]
\[
F(q^\star) = \frac{p - c}{p}.
\]
For a uniform demand distribution on the interval $[a,b]$, its CDF is $F(q) = (q - a)/(b - a)$.
\end{tcolorbox}

Each experimental condition was conducted over 10 independent repetitions with distinct demand sequences to ensure findings are robust across varied demand patterns. Within each repetition, we switch the order of high-margin and low-margin scenarios to evaluate the presentation-order effect, as conjectured in \Cref{hypo:3}.

Through these interactions, we collected a comprehensive dataset including order quantities, complete textual reasoning, and interaction logs. This mixed-method dataset is designed to first quantitatively identify and track decision-making tendencies over time (\textit{e.g.}, learning trajectories and bias types), and then qualitatively analyze the underlying cognitive processes behind these observed behaviors to understand their causes, directly aligning with our research hypotheses.

\subsection{Data Analysis Methods}

Understanding LLM decision making demands a sophisticated analytical framework that can capture both what models decide and how they think about those decisions. We developed a multi-dimensional approach that combines rigorous quantitative metrics with deep qualitative analysis, allowing us to trace the cognitive mechanisms underlying observed behaviors.

\vspace{3mm}
\noindent {\bf Quantitative Methods}
\vspace{3mm}

\noindent Our quantitative framework centers on capturing the multifaceted nature of decision-making biases and learning dynamics. It serves as a comprehensive diagnostic toolkit that reveals different aspects of how LLMs deviate from optimal behavior.

\begin{itemize}
    \item \textbf{Core Bias Measures}. $\;$ Let $q$ denote the LLM order quantity and $q^\star$ the mathematically optimal quantity. The order bias is defined as:
\end{itemize}
\[
\text{OB} = q - q^\star.
\]

To enable fair comparisons across conditions, we standardize this metric as the normalized bias:
\[
\text{NB} = \frac{q - q^\star}{q^\star} \times 100\%,
\]

\begin{itemize}
    \item \textbf{Anchoring Analysis}. $\;$ Mean-anchoring, a key human bias, involves anchoring on expected demand with insufficient adjustment toward optimality. We quantify this using the mean adjustment score (MAS), measuring the extent of adjustment from the cognitive anchor (denoted by $A$ and typically the demand mean) to $q^\star$.
        \begin{itemize}
        \item For high-margin scenarios:
        \begin{equation}\label{eq:mas_high}
            \text{MAS}_{\text{high}} = \frac{q^\star - A}{q - A}.
        \end{equation}
        
        \item For low-margin scenarios:
        \begin{equation}\label{eq:mas_low}
            \text{MAS}_{\text{low}} = \frac{A - q^\star}{A - q}.
        \end{equation}
        \end{itemize}
    A score of 1 indicates full adjustment, whereas 0 indicates no adjustment:
\end{itemize}

\begin{itemize}
    \item \textbf{Economic Impact}. $\;$ We quantify the performance affected by biases via profit efficiency, defined as the ratio of expected profit under the optimal decision to that under the actual decision:
\end{itemize}

\[
\text{PE} = \frac{\mathbb{E}_F[\pi(q^\star, \tilde{d})]}{\mathbb{E}_F[\pi(q, \tilde{d})]} \times 100\%.
\]

\begin{itemize}
    \item \textbf{Dynamic Behavior Analysis}. $\;$ To assess sequential dynamics that are critical for testing presentation-order effect (\Cref{hypo:3}) and overreaction to recent demand signals (\Cref{hypo:4}), we analyze adjustments relative to prior forecast errors. For each round, we define:
    \[
    \Delta t = q_t - q_{t-1} ~~~{\rm and}~~~ \epsilon_{t-1} = d_{t-1} - q_{t-1}.
    \]
    Adjustments are categorized by the sign of $\Delta t \cdot \epsilon_{t-1}$.

    \begin{itemize}
        \item Toward Demand: $\Delta t \cdot \epsilon_{t-1} > 0$, so that the order adjusts in the direction of the prior error (\textit{e.g.}, increasing after a shortage).
        \item Away from Demand: $\Delta t \cdot \epsilon_{t-1} < 0$, so that the order adjusts oppositely.
        \item No Change: $\epsilon_{t-1} = 0$, so that the order remains unchanged.
    \end{itemize}
\end{itemize}

We also measure adjustment magnitude (AM) to quantify the intensity of changes:
\begin{equation} \label{eq:am}
\text{AM} = |q_t - q_{t-1}|.
\end{equation}

To evaluate learning over time (\Cref{hypo:5}), we compute two slopes by regressing against the experimental round $t$. The bias trend slope tracks reductions in order bias magnitude:
\begin{equation}\label{eq:bias_slope}
|q_t - q^\star| \sim t;
\end{equation}
while the profit efficiency trend slope monitors improvements in economic performance:
\begin{equation}\label{eq:pe_slope}
\text{PE}_t \sim t.
\end{equation}

Additionally, we assess adaptation through the change in error responsiveness $\Delta R^2$, derived from the change in variance explained $R^2$ in error-responsiveness regressions between the early stage (rounds~1–7) and the late stage (rounds~8–15).

\vspace{3mm}
\noindent {\bf Qualitative Methods}
\vspace{3mm}

\noindent While quantitative metrics reveal decision outcomes, qualitative analysis uncovers the reasoning processes behind them. We developed a systematic approach to trace the cognitive process from problem interpretation to justification, enabling a detailed reconstruction of LLMs' decision-making pathway.

Our analysis examines each decision through four sequential perspectives. First, information representation captures how the model interprets the newsvendor task---identifying which aspects of profit margins, demand distributions, or cost structures are emphasized or overlooked, thereby revealing its internal problem framing. Second, heuristic activation identifies the cognitive shortcuts at play, such as mean-anchoring, overreaction to recent demand signals, or simplified cost–benefit reasoning, shedding light on the strategies driving the decision. Third, rationale generation focuses on how the model justifies its choice, examining linguistic patterns, rhetorical framing, and possible misconceptions (\textit{e.g.}, invoking risk in risk-neutral contexts) to reveal its justification framework. Finally, feedback integration considers how prior outcomes influence subsequent decisions in multi-round settings, tracing how the model explains its adjustments---or maintains its course---thereby uncovering its learning capacity, adaptive behavior, and persistence of cognitive tendencies.

This structured approach moves beyond simple outcome classification to provide a nuanced understanding of cognitive pathways underlying LLM decision making.

\vspace{3mm}
\noindent {\bf Explanatory Mechanisms}
\vspace{3mm}

\noindent To interpret our findings, we propose four explanatory mechanisms that offer a structured framework linking LLM architecture to the cognitive processes behind observed operational behaviors.

\begin{itemize}
    \item \textit{Dual Representation Mechanism}. $\;$ Drawing parallels to the System 1/System 2 framework in human psychology (\citealp{kahneman2011thinking}), this mechanism posits that LLMs operate with two concurrent processing systems. An analytical module generates optimal solutions when explicit constraints are provided, while a heuristic module dominates in ambiguous contexts, leading to competing influences of heuristic and analytical processes. This accounts for instances where models compute correct solutions but exhibit biased ordering tendencies.

    \item \textit{Attentional Anchoring and Recency Mechanism}. $\;$ This mechanism suggests that LLMs’ Transformer-based attention mechanisms (\citealp{vaswani2017attention}) prioritize early or recent inputs, resulting in sequential processing biases. These biases manifest as anchoring on initial contextual cues (\textit{i.e.}, primacy effect) or overreaction to recent demand signals (\textit{i.e.}, recency effect), creating temporal inertia in ordering decisions. This aligns with human cognitive findings on anchoring (\citealp{tversky1974judgment}) and recency biases (\citealp{hogarth1992order}).

    \item \textit{Corpus-Based Heuristics Mechanism}. $\;$ This mechanism recognizes that LLMs adopt statistical patterns from their training corpora, which act as strong priors shaping decision making. These priors, analogous to Bayesian inference under uncertainty (\citealp{chater2010bayesian}), influence ordering tendencies by reinforcing patterns from the training data.

    \item \textit{Semantic Interference Mechanism}. $\;$ This mechanism posits that natural language framing can override mathematical reasoning in LLMs, similar to the Stroop Effect, where semantic content interferes with task performance (\citealp{macleod1991half}). LLMs may prioritize linguistically salient but mathematically suboptimal responses, particularly in contexts where narrative framing conflicts with optimal strategies.
\end{itemize}

These mechanisms operate synergistically to explain our empirical findings. Table~\ref{tab:hypo-theory-map} maps each hypothesis to key findings, primary explanatory mechanisms, and architectural insights, summarizing a roadmap and the key findings from our analysis of LLM decision making in the newsvendor problem.

\begin{table}[tb]
\centering
\footnotesize
\caption{\protect\mbox{Mapping of Hypotheses to Key Findings, Explanatory Mechanisms, and Architectural Insights}}
\label{tab:hypo-theory-map}
\renewcommand{\arraystretch}{1.0}

\resizebox{\textwidth}{!}{%
\begin{tabular}{
>{\centering\arraybackslash}m{0.15\textwidth} 
>{\arraybackslash}m{0.23\textwidth} 
>{\raggedright}m{0.25\textwidth} 
>{\arraybackslash}m{0.37\textwidth} 
}
\toprule
\textbf{Hypothesis} & 
\makecell[c]{\textbf{Key Finding}} & 
\makecell[c]{\textbf{Explanatory Mechanism}} & 
\makecell[c]{\textbf{Architectural} \textbf{Insight}} \\
\midrule
\makecell[c]{\Cref{hypo:1} \\ (\Cref{sec：H1})} & 
Sys\-tem\-atic order\-ing bias: under\-order\-ing in high-mar\-gin and over\-order\-ing in low-mar\-gin scen\-ar\-io, with ampli\-fi\-ca\-tion be\-yond hu\-man lev\-els. 
& 
\textit{Dual Representation Mechanism}\newline 

\textit{Corpus-Based Heuristics Mechanism} & 
De\-vi\-a\-tion in\-ten\-si\-ty re\-flects the bal\-ance be\-tween heu\-ris\-tic (\textit{e.g.}, mean-an\-chor\-ing) and an\-a\-lyt\-i\-cal pro\-cess\-es. LLaMA-8B's pro\-nounced de\-vi\-a\-tions in\-di\-cate com\-pu\-ta\-tion\-al con\-straints lim\-it\-ing an\-a\-lyt\-i\-cal pro\-cess\-ing. 
\\
\midrule
\makecell[c]{\Cref{hypo:2} \\ (\Cref{sec:H2})} & 
Per\-sis\-tent de\-vi\-a\-tions in risk-neu\-tral en\-vi\-ron\-ments with gua\-ran\-teed pos\-i\-tive prof\-its. 
& 
\textit{Dual Representation Mechanism}\newline

\textit{Semantic Interference Mechanism} & 
De\-vi\-a\-tions are in\-trin\-sic, not risk-driv\-en. GPT-4's ar\-chi\-tec\-ture is sus\-cep\-ti\-ble to se\-man\-tic in\-ter\-fer\-ence, mis\-in\-ter\-pret\-ing risk in risk-neu\-tral set\-tings, while LLaMA-8B ex\-hib\-its fun\-da\-men\-tal heu\-ris\-tic er\-rors. 
\\
\midrule
\makecell[c]{\Cref{hypo:3} \\ (\Cref{sec:H3})} & 
Path de\-pen\-dence: order\-ing an\-chored to in\-i\-tial prof\-it mar\-gin se\-quenc\-es. 
& 
\textit{Attentional Anchoring and Recency Mechanism}\newline

\textit{Dual Representation Mechanism} & 
Self-at\-ten\-tion mech\-a\-nisms' sen\-si\-tiv\-i\-ty to ear\-ly cues cre\-ates de\-ci\-sion in\-er\-tia, a core fea\-ture of Trans\-form\-er-based se\-quen\-tial pro\-cess\-ing. 
\\
\midrule
\makecell[c]{\Cref{hypo:4} \\ (\Cref{sec:H4})} & 
Over\-re\-ac\-tion to re\-cent de\-mand sig\-nals, ad\-just\-ing or\-ders based on lat\-est out\-comes. 
& 
\textit{Attentional Anchoring and Recency Mechanism} & 
The ar\-chi\-tec\-tur\-al ten\-den\-cy to over\-weight re\-cent in\-puts drives over\-re\-ac\-tion to re\-cent de\-mand sig\-nals, re\-flect\-ing re\-cen\-cy bias in\-her\-ent in at\-ten\-tion mech\-a\-nisms. 
\\
\midrule
\makecell[c]{\Cref{hypo:5} \\ (\Cref{sec:H5})} & 
Di\-ver\-gent learn\-ing tra\-jec\-to\-ries: mod\-els ex\-hib\-it dis\-tinct de\-ci\-sion-mak\-ing ten\-den\-cies. 
& 
\textit{Dual Representation Mechanism} & 
Ar\-chi\-tec\-tur\-al trade-offs shape learn\-ing: GPT-4's com\-plex\-i\-ty leads to over\-an\-al\-y\-sis, GPT-4o's ef\-fi\-cien\-cy sup\-ports ra\-tion\-al ex\-e\-cu\-tion, and LLaMA-8B's lim\-i\-ta\-tions hin\-der ef\-fec\-tive learn\-ing. 
\\
\bottomrule
\end{tabular}%
}
\end{table}

The framework we design and develop advances our empirical investigation beyond basic bias identification, offering a systematic exploration of AI cognition in the newsvendor problem. By linking specific findings to explanatory mechanisms and architectural features, we elucidate the underlying causes of LLM decision-making behaviors, providing insights for designing more robust AI systems for high-stakes operational decisions.

\section{Results and Analysis}\label{sec: results}

This section presents the empirical findings from our experiments. We systematically analyze the results to test our five hypotheses, for each of which, we combine quantitative evidence with qualitative analysis of the models' reasoning.

\subsection{Systematic Ordering Bias}\label{sec：H1}

Our \cref{hypo:1} posited that LLMs would replicate the classic underordering in high-margin scenarios and overordering in low-margin scenarios as observed in human decision makers. Our findings confirm this bias and reveal a notable extension: LLMs consistently replicate and occasionally intensify these deviations beyond human benchmarks.

\vspace{3mm}
\noindent {\bf Quantitative Analysis}
\vspace{3mm}

The quantitative results reveal a strong and consistent pattern of ordering bias across the tested LLMs, largely confirming \Cref{hypo:1}. However, they also reveal important nuances, demonstrating that the nature and magnitude of this bias are contingent upon the underlying demand structure.

As summarized in Table~\ref{tab:bias_table}, under the uniform and normal distributions, all three LLMs consistently replicated the classic ``pull-to-center'' effect. They exhibited significant underordering in high-margin scenarios (\textit{e.g.}, an average deviation of -56.29 for the uniform distribution) and pronounced overordering in low-margin scenarios (\textit{e.g.}, an average deviation of +45.36 for the normal distribution). This behavior closely mirrors the patterns observed in human decision makers. However, the lognormal distribution introduced a key exception to this rule. While LLM models still overordered in the low-margin scenarios, they did not uniformly underorder in the high-margin scenarios. In fact, both GPT-4 and GPT-4o exhibited slight overordering, with mean deviations of +3.08 and +0.58, respectively. This finding is critical, as it suggests that the direction and magnitude of the bias are not absolute but are modulated by the skewness and properties of the demand distribution.

\begin{table}[tb]
\centering
\footnotesize
\caption{Systematic Ordering Biases, measured by $q - q^\star$, in Experiment~1}
\label{tab:bias_table}
\renewcommand{\arraystretch}{1.2}
\begin{tabular}{
>{\raggedright\arraybackslash}p{1.8cm}
>{\raggedright\arraybackslash}p{1.8cm}
>{\centering\arraybackslash}p{2cm}
>{\centering\arraybackslash}p{1.8cm}
>{\centering\arraybackslash}p{1.5cm}
>{\centering\arraybackslash}p{2cm}
>{\centering\arraybackslash}p{1.8cm}
>{\centering\arraybackslash}p{1.5cm}
}
\toprule
\multirow{2}{*}{\makecell[l]{Distribution}} & 
\multirow{2}{*}{\makecell[l]{Model}} & 
\multicolumn{2}{c}{\makecell[c]{High-Margin Scenario}} & 
\multirow{2}{*}{\makecell[c]{Deviation}} & 
\multicolumn{2}{c}{\makecell[c]{Low-Margin Scenario}} & 
\multirow{2}{*}{\makecell[c]{Deviation}} \\
\cmidrule{3-4} \cmidrule{6-7}
& & \makecell[c]{Mean Order $q$} & \makecell[c]{Optimal $q^\star$} & & \makecell[c]{Mean Order $q$} & \makecell[c]{Optimal $q^\star$} & \\
\midrule
\multirow{4}{*}{Uniform} 
& GPT-4     & 182.42 & \multirow{4}{*}{\underline{225}} & $-42.58$ & 175.25 & \multirow{4}{*}{\underline{75}} & $+100.25$ \\
& GPT-4o    & 176.03 &                                 & $-48.97$ & 168.89 &                                 & $+93.89$  \\
& LLaMA-8B  & 147.72 &                                 & $-77.28$ & 158.45 &                                 & $+83.45$  \\
& Humans\textsuperscript{1} & 176.83 &                 & $-48.17$ & 134.06 &                                 & $+59.06$  \\
\cmidrule{1-8}
\multirow{3}{*}{Normal} 
& GPT-4     & 181.07 & \multirow{3}{*}{\underline{184}} & $-2.93$  & 175.58 & \multirow{3}{*}{\underline{117}} & $+58.58$  \\
& GPT-4o    & 169.55 &                                 & $-14.45$ & 157.88 &                                 & $+40.88$  \\
& LLaMA-8B  & 154.38 &                                 & $-29.62$ & 153.61 &                                 & $+36.61$  \\
\cmidrule{1-8}
\multirow{3}{*}{Lognormal} 
& GPT-4     & 168.08 & \multirow{3}{*}{\underline{165}} & $+3.08$  & 163.50 & \multirow{3}{*}{\underline{135}} & $+28.50$  \\
& GPT-4o    & 165.58 &                                 & $+0.58$  & 138.88 &                                 & $+3.88$   \\
& LLaMA-8B  & 144.87 &                                 & $-20.13$ & 146.55 &                                 & $+11.55$  \\
\bottomrule
\multicolumn{8}{p{0.95\textwidth}}{\textsuperscript{1} Data source: \cite{schweitzer2000decision}, experiment~1.}
\end{tabular}
\end{table}

\begin{figure}[tb]
    \centering
    \includegraphics[width=1\linewidth]{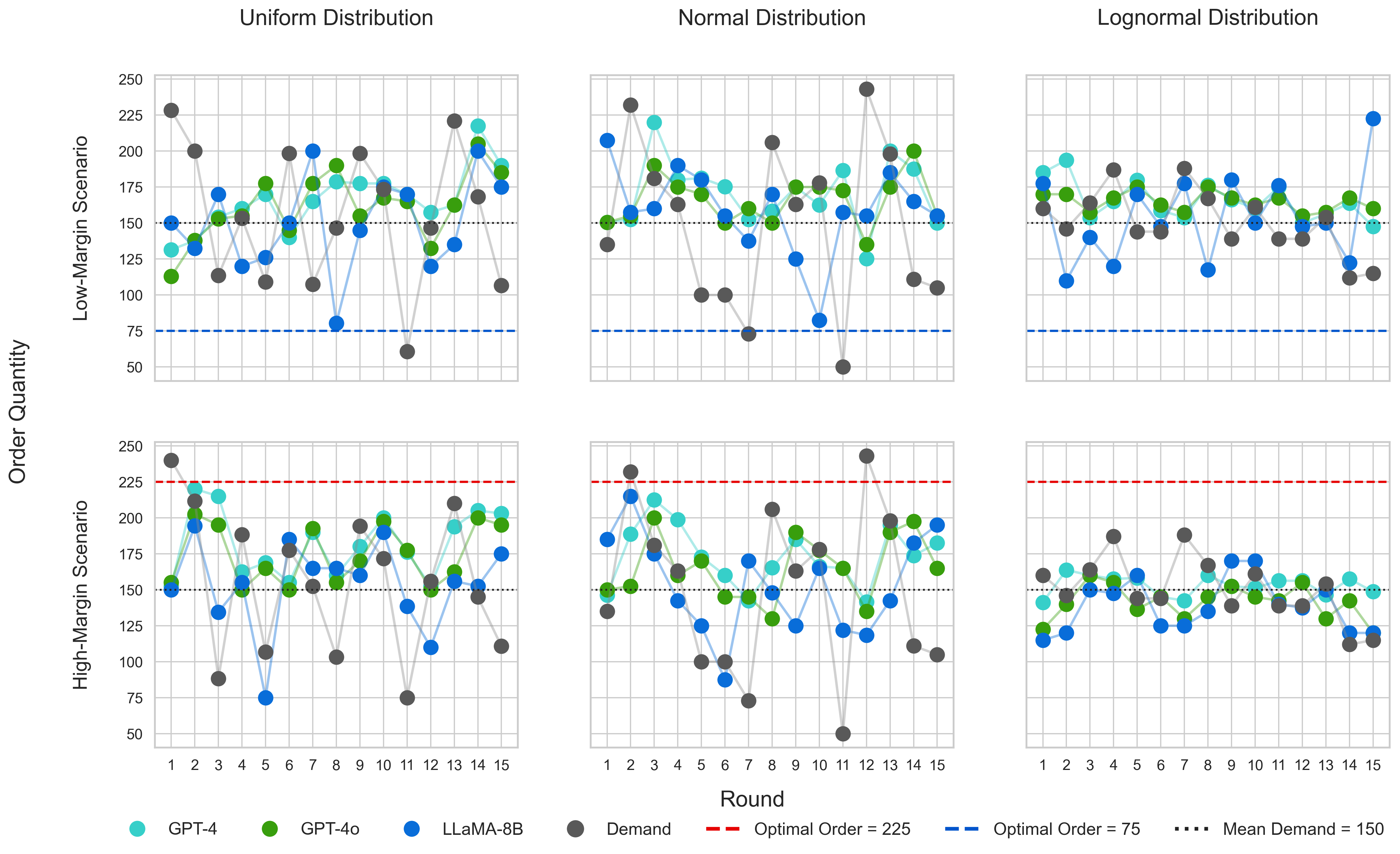}
    \caption{LLM Ordering Trajectories under Different Demand and Profit Scenarios}
    \label{fig:order_trajectory}
\end{figure}

Figure~\ref{fig:order_trajectory} visually corroborates these findings. The ordering trajectories for all models consistently cluster around the fixed mean demand of 150 (the dotted line). This clustering provides strong visual evidence that mean-anchoring is a primary cognitive mechanism, as the models are persistently drawn to the center of the demand range, often failing to adjust sufficiently toward the mathematically optimal quantity---which itself varies significantly with the demand distribution (\textit{e.g.}, from 225 in the uniform high-margin scenario down to 165 for the lognormal).

Notably, these biases are not just replicated but often amplified beyond human levels. For example, in the low-margin uniform condition, GPT-4’s overordering deviation (+100.25) exceeded the human benchmark (+59.06) by~70\%, suggesting that training on human-generated text may reinforce and intensify heuristic tendencies.

Finally, we quantified mean-anchoring using MAS, as defined in~\eqref{eq:mas_high} and~\eqref{eq:mas_low}, with results shown in Table~\ref{tab:qn_table}. The scores, often far from the optimal value of 1, confirm that models made insufficient adjustments to mean demand. The prevalence of negative MAS scores in low-margin scenarios is especially telling, indicating models not only failed to move toward the low optimum but often adjusted in the opposite direction, worsening the bias.

\begin{table}[tb]
\centering
\footnotesize
\caption{MAS for Anchoring Analysis}
\label{tab:qn_table}
\renewcommand{\arraystretch}{1.2}
\begin{tabular}{
>{\raggedright\arraybackslash}p{2.4cm}
>{\raggedright\arraybackslash}p{2.1cm}
>{\centering\arraybackslash}p{1.5cm}
>{\centering\arraybackslash}p{1.5cm}
>{\centering\arraybackslash}p{1.5cm}
>{\centering\arraybackslash}p{1.5cm}
>{\centering\arraybackslash}p{1.5cm}
>{\centering\arraybackslash}p{1.5cm}
}
\toprule
\multirow{2}{*}{\makecell[l]{Condition}} & 
\multirow{2}{*}{\makecell[l]{Model}} & 
\multicolumn{2}{c}{Uniform} & 
\multicolumn{2}{c}{Normal} & 
\multicolumn{2}{c}{Lognormal} \\
\cmidrule(lr){3-4} \cmidrule(lr){5-6} \cmidrule(lr){7-8}
& & 
High Margin & Low Margin & 
High Margin & Low Margin & 
High Margin & Low Margin \\
\midrule
\multirow{4}{*}{\makecell[l]{High-Margin \\ Scenario First}} 
& GPT-4     & 0.41  & $-0.38$ & 1.01 & $-1.28$ & 1.46 & $-1.02$ \\
& GPT-4o    & 0.40  & $-0.38$ & 0.78 & $-0.27$ & 1.34 & 0.74 \\
& LLaMA-8B  & $-0.04$ & $-0.09$ & 0.36 & $-0.36$ & $-0.27$ & 0.42 \\
& Humans\textsuperscript{1} & 0.36 & 0.21 & --- & --- & --- & --- \\
\cmidrule(lr){1-8}
\multirow{4}{*}{\makecell[l]{Low-Margin \\ Scenario First}} 
& GPT-4     & 0.45 & $-0.29$ & 0.84 & $-0.69$ & 0.95 & $-0.78$ \\
& GPT-4o    & 0.29 & $-0.10$ & 0.35 & $-0.17$ & 0.73 & $-0.51$ \\
& LLaMA-8B  & $-0.01$ & $-0.13$ & 0.05 & $-0.12$ & $-0.41$ & 0.03 \\
& Humans\textsuperscript{1} & 0.24 & $-0.27$ & --- & --- & --- & --- \\
\bottomrule
\multicolumn{8}{p{0.95\textwidth}}{\textsuperscript{1} Data source: \cite{schweitzer2000decision}, experiment~1.}
\end{tabular}
\end{table}

\newpage
\vspace{3mm}
\noindent {\bf Qualitative Analysis}
\vspace{3mm}

The models’ textual rationales reveal the cognitive processes driving these quantitative deviations. LLMs frequently substitute heuristic-based reasoning for precise mathematical optimization, evident in their use of risk-oriented language. In high-margin scenarios, models emphasize avoiding excess inventory, despite mathematical incentives for aggressive ordering, whereas in low-margin scenarios, they prioritize capturing demand, justifying overordering by highlighting potential lost sales.

\begin{figure}[tb]
    \centering
    \includegraphics[width=0.7\textwidth]{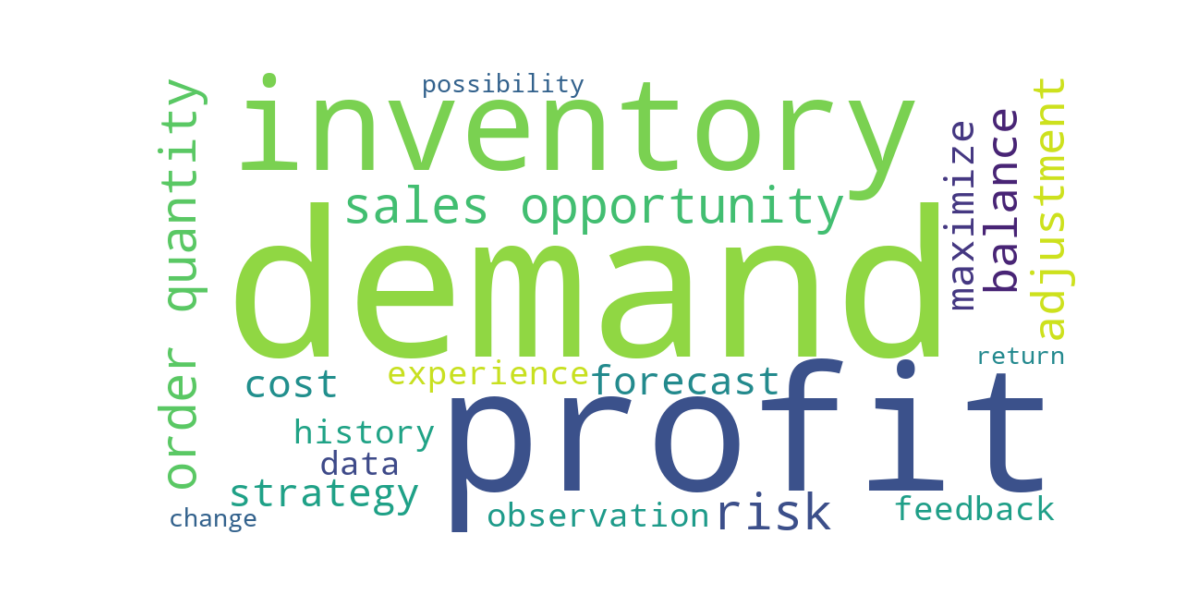}
    \caption{Word Cloud of Decision Rationales}
    \label{fig:bias_wordcloud}
\end{figure}

A word cloud analysis (Figure~\ref{fig:bias_wordcloud}) identifies ``balance'' as a dominant theme, often overriding mathematical logic. For example, GPT-4o, after observing high demand, ordered 230 units (optimum: 75), stating its aim was to ``balance profit maximization with inventory risk minimization''. Notably, some models correctly calculated optimal quantities within their reasoning but adjusted them based on heuristic considerations, such as perceived risk or demand expectations. This supports the \textit{Dual Representation Mechanism}, where heuristic processes compete with analytical reasoning, analogous to the System 1/System 2 framework in human cognition (\citealp{kahneman2011thinking}).

These findings robustly support \Cref{hypo:1} and align with our theoretical framework. The \textit{Dual Representation Mechanism} explains the interplay between analytical competence and heuristic-driven deviations, while the \textit{Corpus-Based Heuristics Mechanism} accounts for the reliance on semantically salient concepts like ``balance'' and ``mean demand''. The intensification of deviations underscores critical implications for deploying LLMs in high-stakes inventory management, where amplified biases could lead to significant economic consequences.

\subsection{Bias Persistence in Risk-Neutral Setting}\label{sec:H2}

Having confirmed that LLMs exhibit systematic ordering bias in the newsvendor problem, we address a critical question: Do these deviations stem from implicit risk preferences, or do they reflect intrinsic cognitive processes? This distinction carries significant implications. If deviations vanish in risk-neutral settings, they may represent rational responses to uncertainty. However, if they persist, they indicate fundamental cognitive tendencies embedded in the models’ architectures.

To distinguish these possibilities, we designed a pivotal test. We created a risk-neutral environment by shifting the demand range to [901, 1200] while maintaining the cost structure, ensuring every ordered unit would sell profitably. This setup eliminates financial risk, isolating the cognitive processes that underlie ordering decisions.

\vspace{3mm}
\noindent {\bf Quantitative Analysis}
\vspace{3mm}

The results strongly support  \Cref{hypo:2}, showing that ordering deviations persist in risk-neutral environments. Table~\ref{tab:H2_risk_free} summarizes this behavior for uniform and normal distributions, revealing distinct architectural influences across the models. Lognormal data was excluded due to non-deterministic outputs (see \Cref{sec:H5}).

\begin{table}[tb]
\centering
\footnotesize
\caption{Ordering Behavior in Risk-Neutral Environment (Experiment~3)}
\label{tab:H2_risk_free}
\renewcommand{\arraystretch}{1.2}
\begin{tabular}{
>{\raggedright\arraybackslash}p{2cm}
>{\centering\arraybackslash}p{2.3cm}
>{\centering\arraybackslash}p{2.3cm}
>{\centering\arraybackslash}p{2.3cm}
>{\centering\arraybackslash}p{2.3cm}
}
\toprule
\multirow{2}{*}{\makecell[l]{Model}} & 
\multicolumn{2}{c}{\makecell[c]{Uniform Distribution}} & 
\multicolumn{2}{c}{\makecell[c]{Normal Distribution}} \\
\cmidrule(lr){2-3} \cmidrule(lr){4-5}
& \makecell[c]{High Margin} & \makecell[c]{Low Margin} & \makecell[c]{High Margin} & \makecell[c]{Low Margin} \\
\midrule
GPT-4 & 
\makecell[c]{1111.95 $|$ \underline{1125} \\ ($-1.16\%$)} & 
\makecell[c]{979.58 $|$ \underline{975} \\ ($+0.47\%$)} & 
\makecell[c]{1069.73 $|$ \underline{1084} \\ ($-1.32\%$)} & 
\makecell[c]{1025.21 $|$ \underline{1017} \\ ($+0.81\%$)} \\
GPT-4o & 
\makecell[c]{1126.2 $|$ \underline{1125} \\ ($+0.11\%$)} & 
\makecell[c]{976.00 $|$ \underline{975} \\ ($+0.10\%$)} & 
\makecell[c]{1083.70 $|$ \underline{1084} \\ ($-0.03\%$)} & 
\makecell[c]{1016.53 $|$ \underline{1017} \\ ($-0.05\%$)} \\
LLaMA-8B & 
\makecell[c]{1001.80 $|$ \underline{1125} \\ ($-10.95\%$)} & 
\makecell[c]{1000.58 $|$ \underline{975} \\ ($+2.62\%$)} & 
\makecell[c]{1016.95 $|$ \underline{1084} \\ ($-6.19\%$)} & 
\makecell[c]{1024.15 $|$ \underline{1017} \\ ($+0.70\%$)} \\
\bottomrule
\multicolumn{5}{p{0.8\textwidth}}{\footnotesize{
\scriptsize{\textit{Notes.} Each cell shows the model’s mean order quantity and the optimal quantity $q^\star$. Values in parentheses indicate normalized bias—the percentage deviation from the optimum.}
}}
\end{tabular}
\end{table}

LLaMA-8B provides the clearest evidence that biases can exist independent of risk. Its substantial deviations---up to -10.95\% in the uniform high-margin scenario---occur despite guaranteed profitability and are manifestations of computational limits preventing accurate calculation of the optimal solution.

GPT-4 exhibits a complex pattern, showing biases even when correctly calculating the optimal solution. The model often computes the correct answer but then applies post-calculation adjustments, a form of ``overthinking''. This sophisticated reasoning paradoxically activates human-like heuristic checks that lead it away from the optimal path, even when no risk is present.

GPT-4o, by contrast, demonstrates markedly different behavior. With deviations from just -0.05\% to +0.11\%, it achieves near-perfect normative rationality. Its decision-making process is direct and mathematically focused, strictly adhering to calculated results without the secondary adjustments that characterize GPT-4. This suggests its efficiency-optimized architecture prioritizes the most direct path to a solution, avoiding qualitative reflections that could introduce bias.

\vspace{3mm}
\noindent {\bf Qualitative Analysis}
\vspace{3mm}

The models’ textual rationales elucidate the cognitive mechanisms driving these deviations, revealing three distinct patterns aligned with their architectural designs. First, GPT-4o’s rationales consistently reference quantile targets and demand distributions, adhering to optimal theory without invoking irrelevant risk concepts. This disciplined, formula-driven approach reflects an architecture optimized for rational execution. In stark contrast, the more sophisticated GPT-4 exhibits a paradoxical irrationality. Despite explicit instructions, it frequently employs risk-oriented language (\textit{e.g.}, ``managing risk'') in risk-neutral contexts, supporting the \textit{Semantic Interference Mechanism} where linguistic associations override mathematical logic. Finally, LLaMA-8B’s deviations stem from fundamental computational errors, such as miscalculating critical ratios (\textit{e.g.}, using 0.25 instead of the correct ratio in a high-margin scenario, yielding $q^\star = 150.5 \times 0.25 = 37.625$), highlighting a misalignment between its probabilistic reasoning and instruction adherence due to architectural limitations.

These results robustly support \Cref{hypo:2}, demonstrating that ordering deviations in risk-neutral environments stem from intrinsic cognitive processes: computational constraints for LLaMA-8B, semantic interference for GPT-4, and the efficacy of a streamlined architecture for GPT-4o. The findings challenge the notion that greater model complexity guarantees superior decisions, suggesting overanalysis can exacerbate biases. This has critical implications for deploying LLMs in high-stakes inventory management, where architectural design significantly influences decision-making efficacy.

\subsection{Presentation-Order Effect} \label{sec:H3}

Having established that LLMs exhibit systematic ordering bias in the newsvendor problem, even in risk-neutral environments, we now examine a key dynamic factor: the influence of initial profit margin conditions on subsequent decision making. Our \Cref{hypo:3} posits that initial conditions establish enduring decision-making frameworks that influence subsequent decisions, even when conditions change significantly. This is critical, as real-world operations involve sequential experiences, and this sequential dependency would mean that LLMs exhibit human-like decision-making processes.

\vspace{3mm}
\noindent {\bf Quantitative Analysis}
\vspace{3mm}

\begin{figure}[tb]
    \centering
    \includegraphics[width=\textwidth]{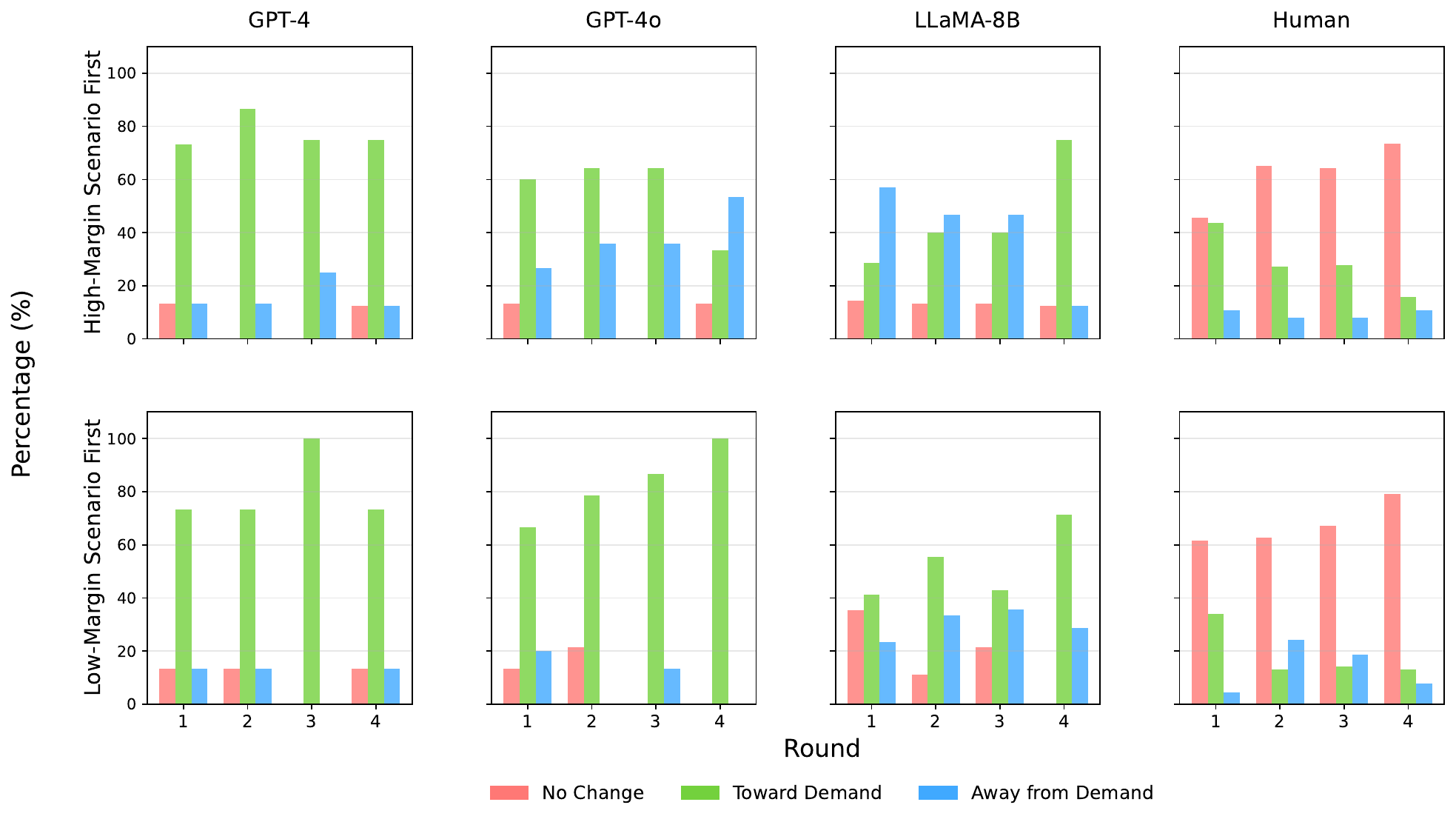}
    \caption{Adjustment Behavior Over Rounds under Uniform Demand Distribution}
    \label{fig:adjustment_uniform}
\end{figure}

To test \Cref{hypo:3} we analyzed round-to-round order adjustments under two scenario sequences: ``High-Margin Scenario First'' and ``Low-Margin Scenario First''. Adjustments were categorized as ``No Change'', ``Toward Demand'', or ``Away from Demand'', reflecting how initial conditions shape subsequent decision making. Figure~\ref{fig:adjustment_uniform}(uniform distribution) shows robust evidence of a presentation-order effect: initial profit margin scenarios establish decision-making frameworks that persist even after conditions change.

When GPT-4 first encountered the high-margin scenario, it exhibited ``Toward Demand'' adjustments 77.5\% of the time during the first four rounds. After transitioning to the low-margin scenario, this rate was still 80.0\% in the early phase of the new condition. If models were purely reactive, we would expect sharp shifts in adjustment patterns; instead, persistence suggests early experiences anchor later behavior. Reversing the sequence reveals a similar pattern: starting with low-margin produces different later responses in the high-margin phase compared to starting with high-margin, indicating systematic path dependence rather than delayed learning.

\begin{figure}[tb]
  \centering
  \includegraphics[height=9cm]{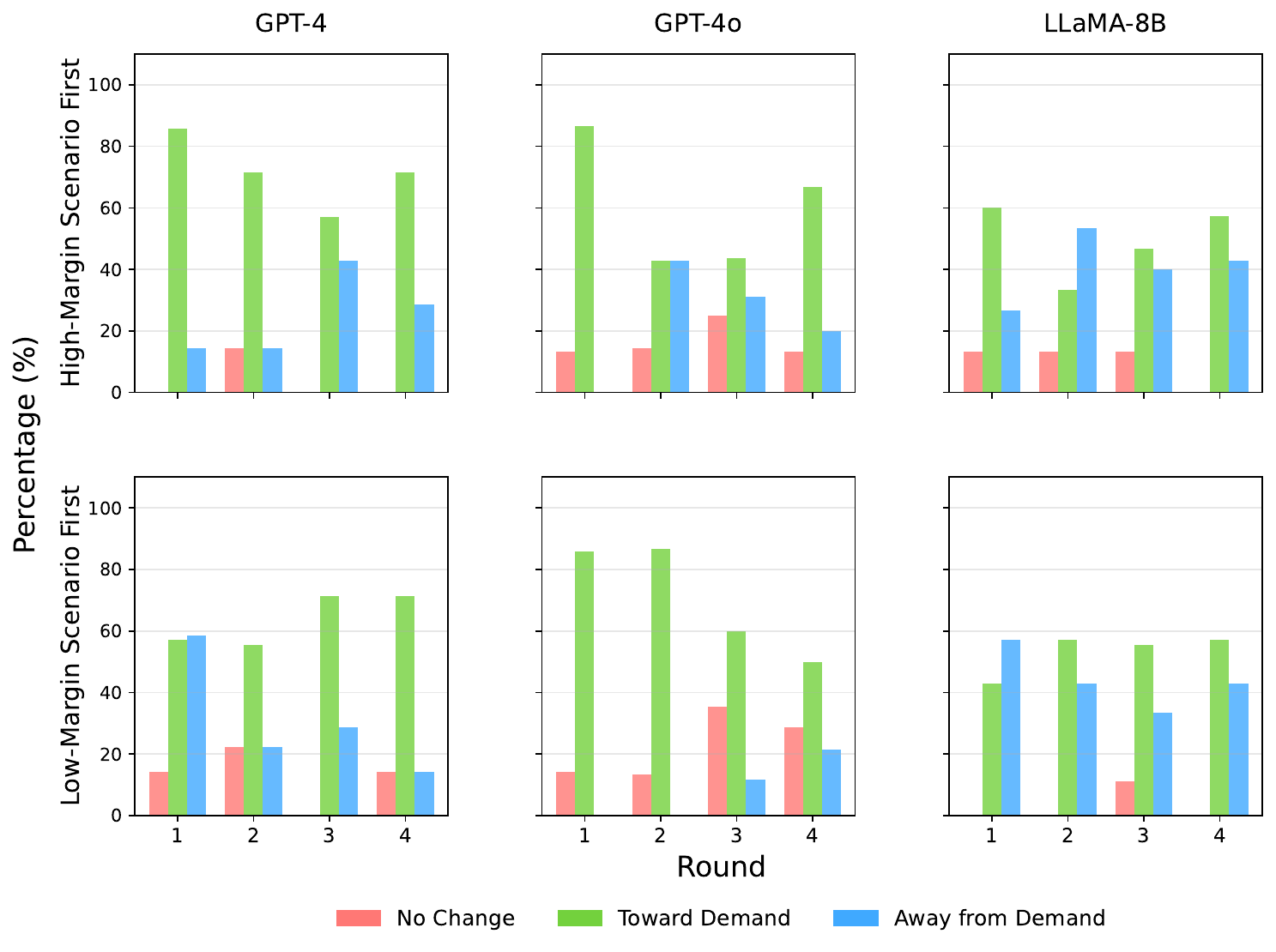}
  \caption{Adjustment Behavior Over Rounds under Normal Demand Distribution}
  \label{fig:adjustment_normal}
\end{figure}

\begin{figure}[tb]
  \centering
  \includegraphics[height=9cm]{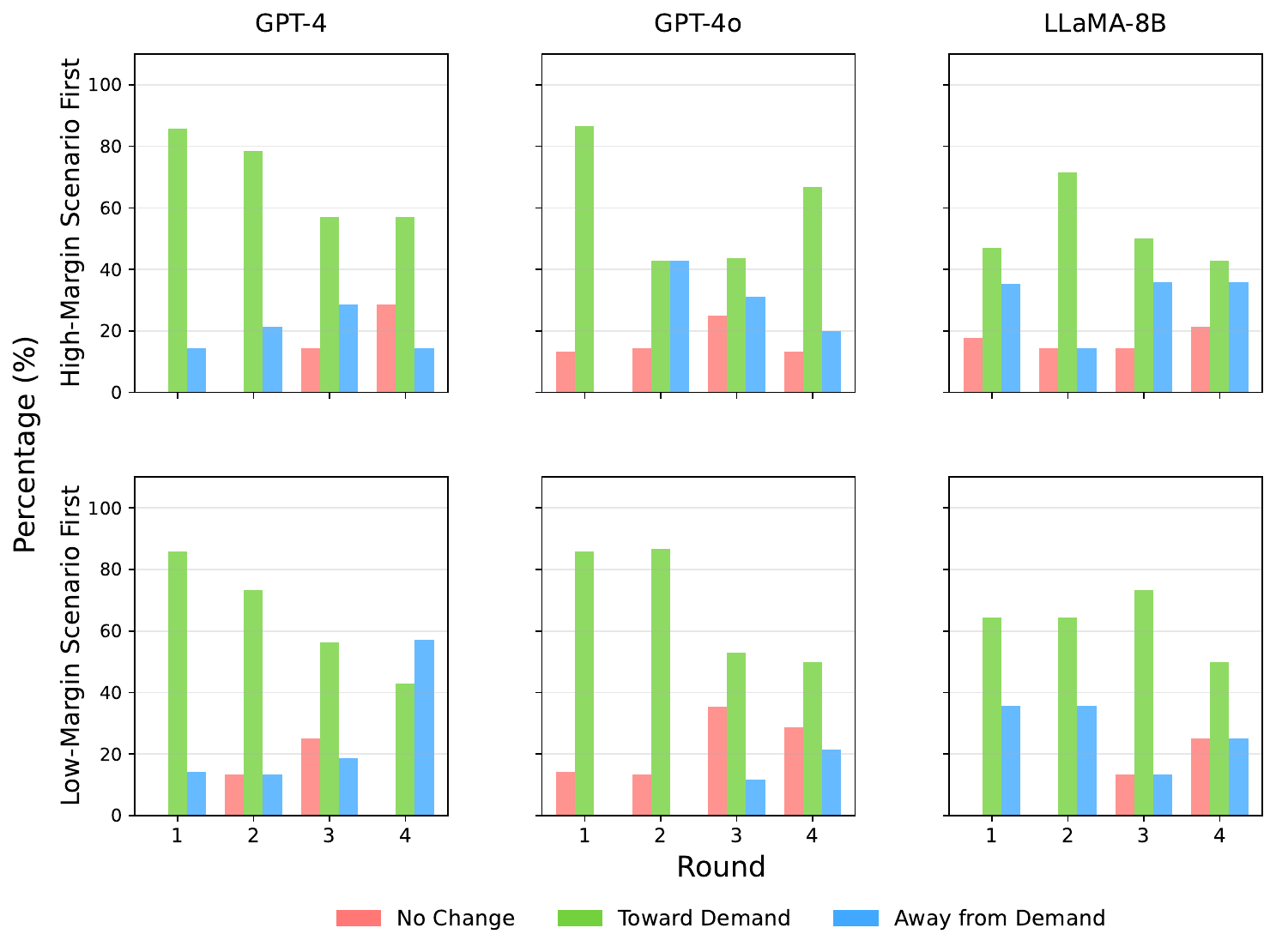}
  \caption{Adjustment Behavior Over Rounds under Lognormal Demand Distribution}
  \label{fig:adjustment_lognormal}
\end{figure}

These effects are consistent across demand distributions, though their magnitude varies. The symmetric normal distribution (Figure~\ref{fig:adjustment_normal}) shows slightly attenuated effects, while the skewed lognormal distribution (Figure~\ref{fig:adjustment_lognormal}) moderates but does not eliminate anchoring. For example, under the lognormal distribution, GPT-4o’s ``Toward Demand'' rate dropped from 85.7\% to 42.9\% after switching profit scenarios, yet still reflected the influence of initial conditions. The fact that these patterns appear across models with different architectures suggests a fundamental feature of sequential processing rather than an architectural artifact.

This persistence aligns with the \textit{Attentional Anchoring and Recency Mechanism}, which posits that Transformer attention mechanisms weight early contextual cues heavily, creating cognitive inertia that resists updating even when conditions clearly change. Path dependence thus emerges as a core property of Transformer-based sequential decision making.

\vspace{3mm}
\noindent {\bf Qualitative Analysis}
\vspace{3mm}

The quantitative evidence reveals clear path dependence, but examining the models' reasoning reveals the cognitive mechanisms underlying this behavioral inertia. Interestingly, different models exhibit distinct forms of cognitive anchoring that reflect their architectural design, even though all show the same basic pattern of presentation-order effect.

GPT-4o exhibits ``semantic template transfer'', reusing generic justifications across profit scenarios. Phrases like ``balancing overordering and underordering risks'' carry over despite changing incentives, indicating an efficient but rigid heuristic strategy. This suggests that once a decision-making framework is established, GPT-4o applies it consistently to minimize computational effort, limiting condition-specific adaptation.

GPT-4 generates elaborate post-hoc rationalizations to justify decisions anchored to initial conditions. After profit condition shifts, it references historical demand patterns or strategic consistency, reflecting a tendency toward overanalysis. This complex reasoning can perpetuate outdated frameworks, hindering optimal adaptation to new conditions.

LLaMA-8B displays fundamental cognitive rigidity, often failing to acknowledge new profit scenarios in its rationales. This suggests its limited representational capacity to detect or adapt to new incentives, with sequential dependency from computational constraints, not strategic anchoring.

These qualitative patterns---efficient template reuse (GPT-4o), complex anchored reasoning (GPT-4), and computational limits (LLaMA-8B)---illuminate the architectural mechanisms driving the presentation-order effect. They support the\textit{ Dual Representation Mechanism}, where heuristic strategies resist reconfiguration, and the \textit{Attentional Anchoring and Recency Mechanism}, where attention mechanisms overweight early cues. This resilience against adaptation highlights critical constraints in LLM sequential processing, with significant implications for their deployment in dynamic operational environments.

\subsection{Demand-Chasing Behavior}\label{sec:H4}

While our previous analyses revealed how LLMs anchor on long-term historical context, we now turn to a very different but equally important phenomenon: their reactivity to the most recent demand signal. Human decision makers consistently exhibit demand chasing---a tendency to overreact to the latest observed demand rather than considering long-term distributional properties. This behavior has been documented across decades of research, from the \cite{niranjan2023unpacking} studies to recent eye-tracking experiments showing that people disproportionately focus on recent demand realizations while underweighting distributional information.

\Cref{hypo:4} predicts that LLMs will exhibit similar or even amplified demand-chasing behavior. Given their sequential processing architecture and training on human-generated content that likely encodes these patterns, we expected strong reactivity to recent signals. The question was not whether this bias would emerge, but how intensely and in what forms across different model architectures.

\vspace{3mm}
\noindent {\bf Quantitative Analysis}
\vspace{3mm}

To analyze this behavior, we categorized each round’s forecast error into absolute magnitude quartiles (Q1–Q4), from smallest to largest, measuring proportions of ``No Change'', ``Toward Demand'', and ``Away from Demand'' adjustments (Table~\ref{tab:H4_data}). This distinguishes reactions to strong, unambiguous signals (Q4) from weak, noisy signals (Q1). Rationally, large errors should prompt corrective adjustments toward demand, while low-error cases reveal inherent biases and heuristics.

\Cref{tab:H4_data} presents a systematic analysis of how models adjust orders by prior forecast error magnitude, revealing a consistent and notable pattern. As error magnitude rises from Q1 to Q4, all LLMs increasingly adjust orders toward demand, with fewer ``No Change'' responses. Under the uniform distribution (low-margin scenario first), GPT-4o’s ``Toward Demand'' rate rises from 82.5\% to 100.0\%, far exceeding human participants, who keep ``No Change'' above 60\% even with large errors.

This demand-chasing behavior differs by architecture. GPT-4 shows over-reactivity---correcting strongly for large errors but overreacting to small ones, particularly in complex distributions---suggesting semantic over-analysis of noise. GPT-4o is consistent, achieving near-perfect ``Toward Demand'' behavior under high-error conditions while maintaining moderate and stable adjustment magnitudes across all error levels, reflecting an efficiency-optimized model. An exception occurs in the high-margin uniform Q1, where GPT-4o makes predominantly away adjustments (83.3\%), indicating that long-term anchoring can override short-term demand signals.

LLaMA-8B is the most erratic. Its ``Toward Demand'' adjustments lack consistency, and in some Q1 settings, ``Away from Demand'' responses dominate. For example, in low-margin scenarios, its Q1 away rate is 43.8\% (uniform distribution) and 56.3\% (lognormal distribution). This tendency to react incorrectly to small, noisy signals reflects architectural and representational limits.

\begin{table}[tb]
\centering
\footnotesize
\caption{Analysis of Demand-Chasing Direction by Error Quartile}
\label{tab:H4_data}
\renewcommand{\arraystretch}{1.2}
\begin{tabular}{
>{\raggedright\arraybackslash}p{1.8cm}
>{\raggedright\arraybackslash}p{1.8cm}
>{\centering\arraybackslash}p{1.4cm}
>{\centering\arraybackslash}p{1.5cm}
>{\centering\arraybackslash}p{1.5cm}
>{\centering\arraybackslash}p{1.5cm}
>{\centering\arraybackslash}p{1.5cm}
>{\centering\arraybackslash}p{1.5cm}
>{\centering\arraybackslash}p{1.5cm}
}
\toprule
\multirow{2}{*}{\makecell[l]{Distribution}} & 
\multirow{2}{*}{\makecell[l]{Model}} & 
\multirow{2}{*}{\makecell[c]{Error \\ Quartile}} & 
\multicolumn{3}{c}{\makecell[c]{Low-Margin Scenario First}} & 
\multicolumn{3}{c}{\makecell[c]{High-Margin Scenario First}} \\
\cmidrule(lr){4-6} \cmidrule(lr){7-9}
& & & 
\makecell[c]{NoChange} & \makecell[c]{Toward} & \makecell[c]{Away} & 
\makecell[c]{NoChange} & \makecell[c]{Toward} & \makecell[c]{Away} \\
\midrule
\multirow{8}{*}{Uniform} 
& \multirow{2}{*}{GPT-4}  
& Q1 & 16.7\% & 72.2\% & 11.1\% & 10.4\% & 70.8\% & 18.8\% \\
& & Q4 & 7.1\%  & 85.7\% & 7.1\%  & 0.0\%  & 92.9\% & 7.1\%  \\
& \multirow{2}{*}{GPT-4o} 
& Q1 & 11.3\% & 82.5\% & 6.3\%  & 5.6\%  & 11.1\% & 83.3\% \\
& & Q4 & 0.0\%  & 100.0\% & 0.0\% & 8.3\%  & 83.3\% & 8.3\%  \\
& \multirow{2}{*}{LLaMA-8B}  
& Q1 & 18.8\% & 37.5\% & 43.8\% & 5.6\%  & 41.7\% & 52.8\% \\
& & Q4 & 0.0\%  & 69.0\% & 31.0\% & 8.3\%  & 75.0\% & 16.7\% \\
& \multirow{2}{*}{Humans\textsuperscript{1}}  
& Q1 & 63.7\% & 23.1\% & 13.2\% & 62.0\% & 24.8\% & 13.1\% \\
& & Q4 & 71.8\% & 16.5\% & 11.8\% & 58.8\% & 37.5\% & 3.7\%  \\
\cmidrule(lr){1-9}
\multirow{6}{*}{Normal}  
& \multirow{2}{*}{GPT-4}  
& Q1 & 22.2\% & 55.6\% & 22.2\% & 12.5\% & 62.5\% & 25.0\% \\
& & Q4 & 0.0\%  & 100.0\% & 0.0\% & 0.0\%  & 85.7\% & 14.3\% \\
& \multirow{2}{*}{GPT-4o} 
& Q1 & 18.8\% & 43.8\% & 37.5\% & 25.0\% & 43.8\% & 31.3\% \\
& & Q4 & 0.0\%  & 100.0\% & 0.0\% & 0.0\%  & 100.0\% & 0.0\% \\
& \multirow{2}{*}{LLaMA-8B}  
& Q1 & 6.3\%  & 57.6\% & 36.1\% & 6.3\%  & 50.0\% & 43.8\% \\
& & Q4 & 0.0\%  & 78.6\% & 21.4\% & 7.1\%  & 64.3\% & 28.6\% \\
\cmidrule(lr){1-9}
\multirow{6}{*}{Lognormal} 
& \multirow{2}{*}{GPT-4}  
& Q1 & 6.3\%  & 50.0\% & 43.8\% & 10.6\% & 42.2\% & 47.2\% \\
& & Q4 & 0.0\%  & 71.4\% & 28.6\% & 0.0\%  & 84.5\% & 15.5\% \\
& \multirow{2}{*}{GPT-4o} 
& Q1 & 23.6\% & 65.3\% & 11.1\% & 12.5\% & 56.3\% & 31.3\% \\
& & Q4 & 8.3\%  & 84.5\% & 7.1\%  & 14.3\% & 64.3\% & 21.4\% \\
& \multirow{2}{*}{LLaMA-8B}  
& Q1 & 6.3\%  & 37.5\% & 56.3\% & 6.3\%  & 43.8\% & 50.0\% \\
& & Q4 & 0.0\%  & 100.0\% & 0.0\% & 23.8\% & 69.0\% & 7.1\%  \\
\bottomrule
\multicolumn{9}{p{0.92\textwidth}}{\footnotesize{
\textsuperscript{1} Human data from \cite{schweitzer2000decision}. Error Quartiles (Q1–Q4) categorize the absolute magnitude of the prior round's forecast error, from the smallest (Q1) to the largest (Q4).
}}
\end{tabular}
\end{table}

\begin{figure}[tb] 
    \centering
    \includegraphics[width=1\linewidth]{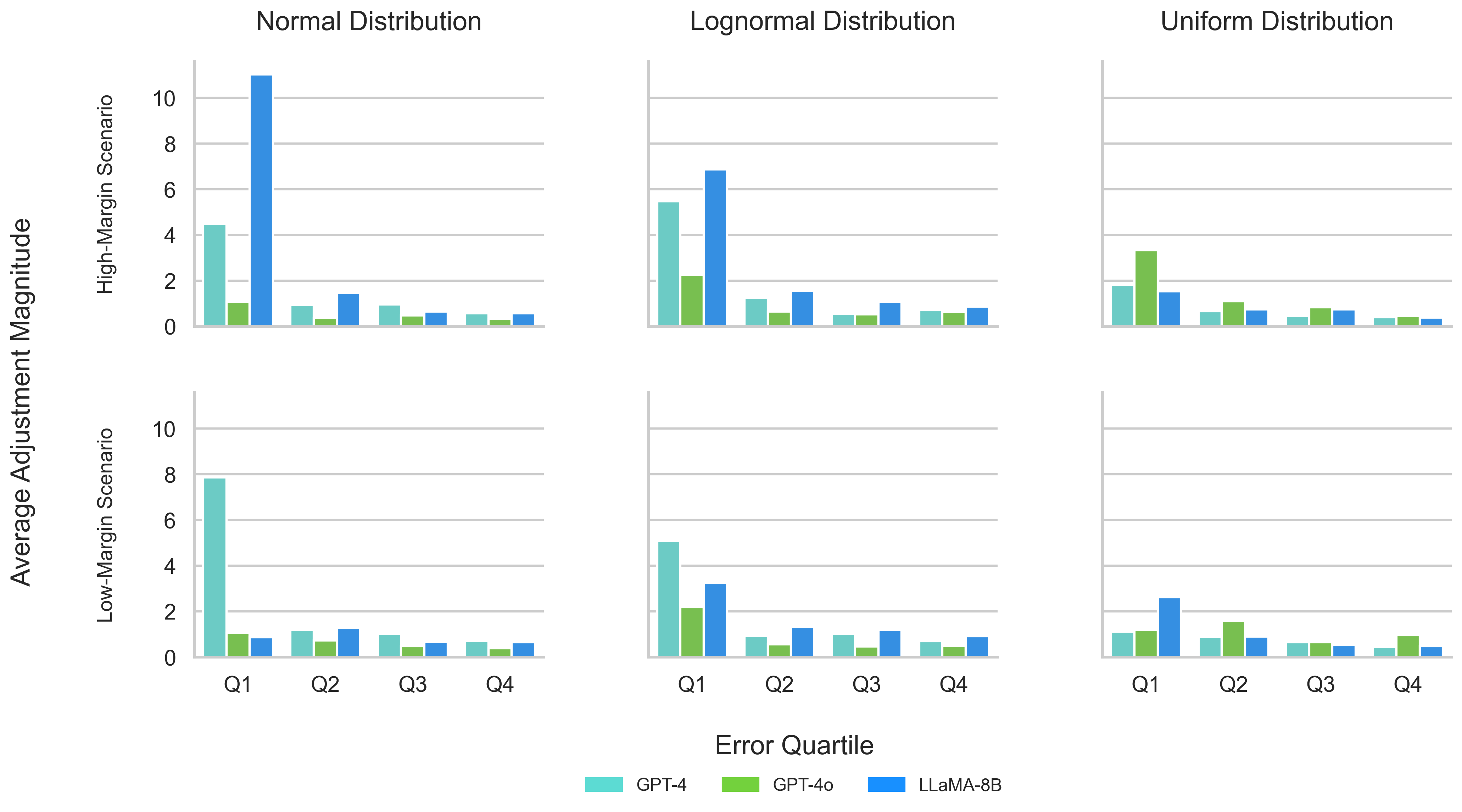} 
    \caption{Magnitude of `Toward Demand' Adjustments by Error Quartile}
    \label{fig:distribution_profit}
\end{figure}

While adjustment direction reveals how models react, adjustment magnitude~\eqref{eq:am} shows their sensitivity calibration---how well they scale reactions to error size. Figure~\ref{fig:distribution_profit} shows that, contrary to rational adjustment, magnitude does not always increase with error size. Both GPT-4 and LLaMA-8B make disproportionately large adjustments under Q1 conditions, especially in the normal and lognormal distributions, indicating poor scaling with error size. In contrast, GPT-4o maintains stable, proportional adjustments that scale appropriately with error magnitude, demonstrating consistent heuristic application.

\vspace{3mm}
\noindent {\bf Qualitative Analysis}
\vspace{3mm}

When we examine the textual reasoning behind these adjustment patterns, the cognitive mechanisms become even clearer. GPT-4o consistently employs a demand-chasing reasoning framework, justifying its adjustments with plausible semantic heuristics like framing the last demand signal as a reasonable anchor and using generic phrases like ``find a balance''. This represents the hallmark of an efficient agent that has identified a simple, effective rule and applies it without costly deliberation.

GPT-4’s reasoning reveals a more complex and paradoxically irrational pattern. Its powerful reasoning capabilities lead it to over-interpret recent demand signals, rationalizing small deviations as evidence of ``new trends'' or ``increased volatility'' that justify its disproportionately large adjustments. This overthinking of noisy, short-term data is a direct manifestation of how sophisticated reasoning can paradoxically lead to less rational decisions.

LLaMA-8B's justifications are best described as erratic and opportunistic, lacking any coherent framework. In one round, it might chase demand, in the next round, it might anchor to the mean, with textual explanations that fail to provide consistent logic. This instability in reasoning directly mirrors the unstable adjustment patterns we observe quantitatively and highlights the brittle nature of heuristics in computationally constrained systems.

These findings provide strong support for our \Cref{hypo:4} while revealing deeper insights. Demand-chasing behavior in LLMs isn't just a replication of human patterns---it's often an amplification that reflects how different AI architectures process sequential information. Our \textit{Attentional Anchoring and Recency Mechanism} explains this phenomenon: Transformer-based models tend to overweight recent inputs, but how this bias manifests depends critically on architectural choices. GPT-4's depth enables over-interpretation, GPT-4o's streamlined design produces stable heuristics, and LLaMA-8B's scale limitations lead to inconsistent responses.

Most significantly, these results demonstrate that LLMs exhibit demand-chasing behavior that often exceeds human levels---with adjustment rates approaching 100\% in high-error conditions compared to human rates that rarely exceed 40\%. This amplification of human biases represents a critical finding for the deployment of AI systems in operational contexts, where such over-reactivity could lead to significant inefficiencies and instabilities.

\subsection{Constraints on Learning from Feedback}\label{sec:H5}

Our prior analyses confirmed that LLMs exhibit systematic ordering bias, overreact to recent demand signals, and display sequential dependency in the newsvendor problem. This leads to a critical question for practical deployment: Can LLMs learn from feedback to mitigate these biases? \Cref{hypo:5} examines whether learning is constrained by architectural limitations. To assess learning capacity, we compared Experiment~1 and Experiment~2, testing the ability to translate theoretical knowledge into performance. The contrast elucidates both the extent and nature of learning in LLMs.

\vspace{3mm}
\noindent {\bf Quantitative Analysis}
\vspace{3mm}

We analyzed the models' adaptation patterns using three core metrics: the Convergence Slope defined in~\eqref{eq:bias_slope} to track bias reduction over time; the Efficiency Slope from defined in~\eqref{eq:pe_slope} to measure improvements in economic performance; and the Change in Error Responsiveness $\Delta R^2$ to assess feedback sensitivity. Table~\ref{tab:llm_learning_exp1} and Table~\ref{tab:llm_learning_exp2} present the results of this analysis for Experiments~1 and~2, respectively.\footnotemark\footnotetext{Data for the lognormal distribution in Experiment~2 is excluded, as models correctly identified the absence of the inverse cumulative distribution function and produced assumption-based outputs unsuitable for quantitative analysis, though these provide qualitative insights.} The results reveal notable variations in adaptation, highlighting architectural influences on learning behaviors across models.

GPT-4o’s behavior aligns with that of an efficient rationalist. It exhibits near-optimal performance with minimal deviation. While its adaptation was limited in Experiment~1 (without the formula), providing explicit guidance in Experiment~2 led to near-perfect rationality. Its convergence slopes approached zero (\textit{e.g.}, +0.003, -0.007; Table~\ref{tab:llm_learning_exp2}), indicating it immediately adopted the optimal solution without further adjustment.

GPT-4, in contrast, displays the paradoxical adaptation pattern of a complex analyst. While it showed some capacity for self-correction in Experiment~1 (\textit{e.g.}, slope of -1.917), formula guidance in Experiment~2 did not guarantee optimality. In fact, its performance sometimes worsened, with a strongly positive convergence slope (+1.191) in the uniform low-margin scenario. This reflects a deep-seated conflict between its analytical knowledge and heuristic processes.

LLaMA-8B’s performance underscores its role as a constrained adaptor. Its adaptation patterns reveal significant architectural limitations, with inconsistent and often counterproductive learning trends across both experiments. It failed to improve consistently, even with explicit formula guidance in Experiment~2, highlighting its fundamental limitations in processing feedback and applying symbolic rules.

\begin{table}[tb]
\centering
\footnotesize
\caption{LLM Learning Metrics in Experiment~1 (without Formula Guidance)}
\label{tab:llm_learning_exp1}
\renewcommand{\arraystretch}{1.2}
\begin{tabular}{
>{\raggedright\arraybackslash}p{2.4cm}
>{\raggedright\arraybackslash}p{2.1cm}
>{\centering\arraybackslash}p{1.6cm}
>{\centering\arraybackslash}p{1.6cm}
>{\centering\arraybackslash}p{1.6cm}
>{\centering\arraybackslash}p{1.6cm}
>{\centering\arraybackslash}p{1.6cm}
>{\centering\arraybackslash}p{1.6cm}
}
\toprule
\multirow{2}{*}{Distribution} & 
\multirow{2}{*}{Model} & 
\multicolumn{3}{c}{High-Margin Scenario} & 
\multicolumn{3}{c}{Low-Margin Scenario} \\
\cmidrule(lr){3-5} \cmidrule(lr){6-8}
& & 
Convergence & Efficiency & $\Delta R^2$ & 
Convergence & Efficiency & $\Delta R^2$ \\
\midrule
\multirow{3}{*}{Uniform} 
& GPT-4     & +0.410 & +0.004 & -0.300 & +0.773 & +0.007 & +0.120 \\
& GPT-4o    & +0.267 & +0.001 & -0.930 & +0.116 & +0.006 & -0.165 \\
& LLaMA-8B  & +0.107 & +0.004 & +0.015 & -0.391 & +0.011 & -0.250 \\
\midrule
\multirow{3}{*}{Normal} 
& GPT-4     & -0.184 & -0.002 & -0.405 & -0.276 & -0.004 & -0.205 \\
& GPT-4o    & +0.019 & -0.001 & -0.245 & +0.007 & -0.002 & -0.025 \\
& LLaMA-8B  & +0.146 & -0.006 & -0.360 & -0.115 & +0.000 & -0.160 \\
\midrule
\multirow{3}{*}{Lognormal} 
& GPT-4     & -1.917 & +0.008 & +0.430 & -0.578 & -0.032 & +0.100 \\
& GPT-4o    & -0.209 & -0.005 & -0.250 & -0.037 & +0.006 & -0.730 \\
& LLaMA-8B  & +0.209 & -0.007 & -0.230 & +0.106 & -0.003 & +0.400 \\
\bottomrule
\end{tabular}
\end{table}

\begin{table}[tb]
\centering
\footnotesize
\caption{LLM Learning Metrics in Experiment~2 (with Formula Guidance)}
\label{tab:llm_learning_exp2}
\renewcommand{\arraystretch}{1.2}
\begin{tabular}{
>{\raggedright\arraybackslash}p{2.4cm}
>{\raggedright\arraybackslash}p{2.1cm}
>{\centering\arraybackslash}p{1.6cm}
>{\centering\arraybackslash}p{1.6cm}
>{\centering\arraybackslash}p{1.6cm}
>{\centering\arraybackslash}p{1.6cm}
>{\centering\arraybackslash}p{1.6cm}
>{\centering\arraybackslash}p{1.6cm}
}
\toprule
\multirow{2}{*}{Distribution} & 
\multirow{2}{*}{Model} & 
\multicolumn{3}{c}{High-Margin Scenario} & 
\multicolumn{3}{c}{Low-Margin Scenario} \\
\cmidrule(lr){3-5} \cmidrule(lr){6-8}
& & 
Convergence & Efficiency & $\Delta R^2$ & 
Convergence & Efficiency & $\Delta R^2$ \\
\midrule
\multirow{3}{*}{Uniform}
& GPT-4     & +0.484 & +0.005 & -0.910 & +1.191 & +0.004 & -0.845 \\
& GPT-4o    & +0.003 & +0.000 & -0.960 & -0.201 & -0.003 & -0.825 \\
& LLaMA-8B  & -0.137 & +0.003 & +0.070 & -0.590 & +0.012 & -0.869 \\
\midrule
\multirow{3}{*}{Normal}
& GPT-4     & +0.072 & +0.002 & -0.538 & -0.151 & +0.002 & +0.329 \\
& GPT-4o    & -0.007 & -0.000 & -0.376 & -0.005 & +0.000 & -0.491 \\
& LLaMA-8B  & +0.161 & -0.009 & -0.181 & -0.111 & +0.006 & -0.031 \\
\bottomrule
\end{tabular}
\end{table}

\vspace{3mm}
\noindent {\bf Qualitative Analysis}
\vspace{3mm}

The models' textual rationales illuminate the cognitive mechanisms driving these patterns of adaptation.GPT-4o's rationales in Experiment~2 are concise and formula-driven, directly applying the optimal rule without extraneous elaboration. This aligns with its near-zero convergence slopes, reflecting a streamlined architecture prioritizing mathematical precision over heuristic adjustments.

GPT-4's rationales reveal a tendency toward overanalysis, layering heuristic judgments atop mathematical results. For instance, in the uniform low-margin scenario of Experiment~2, it stated, ``The formula suggests an order of 225, but I adjust to 210 to mitigate surplus risk'', despite the optimal solution. This reflects a conflict between analytical and heuristic processes, supporting the \textit{Dual Representation Mechanism}, where implicit heuristics override explicit knowledge. In the lognormal condition, GPT-4 recognized the missing inverse CDF and provided hypothetical responses, avoiding flawed numerical outputs. LLaMA-8B's rationales expose computational limitations, often reverting to simplistic heuristics like ``ordering around the demand mean of 150''. In Experiment~2's lognormal condition, it erroneously assumed a normal distribution, highlighting an inability to maintain coherent symbolic reasoning under uncertainty.

These findings support \Cref{hypo:5}, revealing that LLM learning is constrained by architectural design. GPT-4o’s streamlined execution achieves near-optimal performance, GPT-4’s overanalysis introduces deviations, and LLaMA-8B’s limitations hinder effective adaptation. The \textit{Dual Representation Mechanism }explains the tension between analytical and heuristic processes, while the \textit{Corpus-Based Heuristics Mechanism} accounts for reliance on ingrained patterns. These results challenge the assumption that providing optimal strategies ensures optimal outcomes, showing that greater model complexity does not guarantee superior performance. This has critical implications for deploying LLMs in dynamic inventory management, where architectural constraints shape learning efficacy.

\section{Concluding Discussion}\label{sec:discussion}

Our study addressed a fundamental question in artificial intelligence: to what extent do LLMs, as emerging economic agents, replicate or amplify human cognitive biases? Using dynamic experiments based on the newsvendor problem, we found that the issue extends beyond replication. Our key theoretical contribution is the \textit{Paradox of Intelligence}: increased model complexity does not ensure optimality and may even increase susceptibility to heuristic deviations. All tested models consistently reproduced core human biases, such as underordering in high-margin and overordering in low-margin scenarios (\Cref{hypo:1}), even in risk-neutral contexts (\Cref{hypo:2}). These are not random errors but cognitive signatures shaped by model architecture: LLaMA-8B’s computational limits led to unstable heuristics; GPT-4o’s efficiency focus enabled consistent rule-following; GPT-4’s complexity introduced overanalysis that hindered optimality.

We identify four mechanisms driving these biases. The \textit{Semantic Interference Mechanism} shows how linguistic framing can override reasoning, leading models to misapply concepts like risk in risk-free tasks. The \textit{Dual Representation Mechanism} explains tension between analytical and heuristic processes. The \textit{Attentional Anchoring and Recency Mechanism} accounts for systematic biases---such as path dependence and overreaction to recent data---emerging from Transformer architectures (\Cref{hypo:3} and \Cref{hypo:4}). Lastly, the \textit{Corpus-Based Heuristics Mechanism} captures how patterns from training data become embedded priors. Treating LLMs as experimental subjects allows us to validate behavioral theories in AI systems and underscores the need for cognitively grounded evaluation frameworks.

These insights have substantial practical implications. Organizations should pursue context-specific model selection, recognizing that efficiency-optimized models like GPT-4o may outperform more complex alternatives in constrained optimization tasks. The strong effect of providing explicit formulas highlights the value of rule-based prompting to activate analytical reasoning. Most importantly, human oversight is essential: LLMs can amplify human biases by up to 70\%, making human-in-the-loop systems critical in domains like inventory management, where such amplification can lead to major economic consequences.

While this study offers a robust framework, several limitations point to future research. A key direction is to explore human-AI collaborative decision making, examining how various oversight mechanisms can reduce persistent biases. Further, extending this framework to multi-agent economic games could shed light on LLM behavior in competitive and cooperative settings. Developing standardized benchmarks for economic rationality will also be essential for tracking model evolution. Ultimately, as more economic decisions are delegated to AI, the challenge is not just building more powerful models---but designing ones that behave in predictably rational ways, ensuring AI advances support rather than undermine sound economic decision making.

{\newcommand{\custombibsize}{\fontsize{11pt}{13.5pt}\selectfont} 

\bibliographystyle{ormsv080}
\renewcommand*{\bibfont}{\custombibsize}
\bibliography{reference}}


\newpage

\ECSwitch 
\ECHead{\centering E-Companion to \\ ``Large Language Newsvendor: \\ Decision Biases and Cognitive Mechanisms''}

This e-companion provides the exact prompt templates used in our experiments, showing the specific text presented to each LLM.

\section{Template of Base Prompt}

\vspace{7mm}
\begin{tcolorbox}[breakable, colback=blue!5, colframe=blue!40, title=Core Prompt Structure, width=\textwidth, before skip=10pt, after skip=10pt]
\begin{sloppypar}
You are participating in an inventory decision experiment. You need to make inventory ordering decisions for selling ``wodgets'' products.\\

Experiment rules:\\
- Each wodget sells for 12 francs\\
- Each wodget costs \{cost\} francs\\
- Remaining unsold wodgets can be disposed of for 0 francs\\
- \{demand\_description\}\\
- Your goal is to maximize profit\\

\{history\_block\}\\

Here is some information that might be helpful:\\
- If sales exceed your order quantity, you will lose some sales opportunities\\
- If the order quantity exceeds sales, you will need to dispose of the remaining inventory \\
- Please consider the costs of ordering too much and ordering too little\\

\{formula\_block\}\\

Please make an inventory ordering decision: How many wodgets will you order? Please explain your thinking process and decision in detail.
\end{sloppypar}
\end{tcolorbox}

\newpage
\section{Parameter Substitutions}

\vspace{3mm}
\noindent {\bf Description of Demand Distribution}
\vspace{3mm}

\begin{tcolorbox}[breakable, colback=blue!3, colframe=blue!40!black,
  title=Uniform Distribution,
  width=\textwidth, before skip=10pt, after skip=10pt]
demand is uniformly distributed between \{a\} and \{b\}
\end{tcolorbox}

\begin{tcolorbox}[breakable, colback=blue!3, colframe=blue!40!black,
  title=Normal Distribution,
  width=\textwidth, before skip=10pt, after skip=10pt]
demand follows a normal distribution with mean \{mean\} and standard deviation approximately \{std\}, mainly ranging between \{a\} and \{b\}
\end{tcolorbox}

\begin{tcolorbox}[breakable, colback=blue!3, colframe=blue!40!black,
  title=Lognormal Distribution,
  width=\textwidth, before skip=10pt, after skip=10pt]
demand follows a right-skewed distribution (similar to lognormal distribution) with mean approximately \{mean\}, having a longer right tail, ranging between \{a\} and \{b\}
\end{tcolorbox}

\vspace{3mm}
\noindent {\bf History Block (Rounds~2--15)}
\vspace{3mm}

\begin{tcolorbox}[breakable, colback=yellow!10, colframe=yellow!50!black,
  title=Feedback Information,
  width=\textwidth, before skip=10pt, after skip=10pt]
In the previous round:\\
- Your order quantity: \{last\_order\} wodgets\\
- Actual demand: \{last\_demand\} wodgets \\ 
- This round's profit: \{last\_profit\} francs

Current cumulative profit: \{cumulative\_profit\} francs
\end{tcolorbox}

\newpage
\section{Formula Block (Experiment~2 Only)}

\vspace{7mm}
\begin{tcolorbox}[breakable, colback=orange!5, colframe=orange!80, title=Theoretical Guidance, width=\textwidth, before skip=10pt, after skip=10pt]
Note: You have previously learned about the newsvendor problem. In this problem, the profit-maximizing order quantity can be determined by the following formula:

\[
F(q^\star) = \frac{p - c}{p},
\]
where $F(q)$ is the cumulative distribution function, $p$ is the selling price, and $c$ is the cost.

\{distribution\_formula\}
\end{tcolorbox}

\vspace{3mm}
\noindent {\bf Guidance of Distribution-Specific Formula}
\vspace{3mm}

\begin{tcolorbox}[breakable, colback=red!5, colframe=red!40, title=Uniform Distribution Formula, width=\textwidth, before skip=10pt, after skip=10pt]
For uniform distribution $\mathbb{U}[\{a\}, \{b\}]$, $F(q) = (q - \{a\}) / (\{b\} - \{a\})$.
\end{tcolorbox}

\begin{tcolorbox}[breakable, colback=red!5, colframe=red!40, title=Normal Distribution Formula, width=\textwidth, before skip=10pt, after skip=10pt]
For normal distribution $\mathbb{N}(\{\text{mean}\}, \{\text{std}\})$, use standard normal CDF for calculation.
\end{tcolorbox}

\begin{tcolorbox}[breakable, colback=red!5, colframe=red!40, title=Lognormal Distribution Formula, width=\textwidth, before skip=10pt, after skip=10pt]
For a right-skewed distribution, calculate the cumulative probability based on specific distribution characteristics.
\end{tcolorbox}

\newpage
\section{Examples of Complete Prompts}

\vspace{3mm}
\noindent {\bf Experiment~1 (High Margin, Uniform Distribution, Round~1)}
\vspace{3mm}

\begin{tcolorbox}[breakable, colback=gray!5!white, colframe=gray!50!black, title=Actual Prompt Text, width=\textwidth, before skip=10pt, after skip=10pt]
\begin{sloppypar}
You are participating in an inventory decision experiment. You need to make inventory ordering decisions for selling ``wodgets'' products.\\

Experiment rules:\\
- Each wodget sells for 12 francs\\
- Each wodget costs 3 francs\\
- Remaining unsold wodgets can be disposed of for 0 francs\\
- demand is uniformly distributed between 1 and 300\\
- Your goal is to maximize profit\\

Here is some information that might be helpful:\\
- If sales exceed your order quantity, you will lose some sales opportunities\\
- If order quantity exceeds sales, you will need to dispose of remaining inventory\\
- Please consider the costs of ordering too much and ordering too little\\

Please make an inventory ordering decision: How many wodgets will you order? Please explain your thinking process and decision in detail.
\end{sloppypar}
\end{tcolorbox}

\vspace{3mm}
\noindent {\bf Experiment~2 (Low Margin, Normal Distribution, Round~5)}
\vspace{3mm}

\begin{tcolorbox}[breakable, colback=gray!5!white, colframe=gray!50!black, title=Actual Prompt Text with History and Formula, width=\textwidth, before skip=10pt, after skip=10pt]
\begin{sloppypar}
You are participating in an inventory decision experiment. You need to make inventory ordering decisions for selling ``wodgets'' products.\\

Experiment rules:\\
- Each wodget sells for 12 francs\\
- Each wodget costs 9 francs\\
- Remaining unsold wodgets can be disposed of for 0 francs\\
- demand follows a normal distribution with mean 150.5 and standard deviation approximately 49.8, mainly ranging between 1 and 300\\
- Your goal is to maximize profit\\

In the previous round:\\
- Your order quantity: 120 wodgets\\
- Actual demand: 85 wodgets\\
- This round's profit: 255 francs\\

Current cumulative profit: 1450 francs
Here is some information that might be helpful:\\
- If sales exceed your order quantity, you will lose some sales opportunities\\
- If the order quantity exceeds sales, you will need to dispose of the remaining inventory\\
- Please consider the costs of ordering too much and ordering too little\\

Note: You have previously learned about the newsvendor problem. In this problem, the profit-maximizing order quantity can be determined by the following formula:
\[
F(q^\star) = \frac{p - c}{p},
\]
where $F(q)$ is the cumulative distribution function, $p$ is the selling price, and $c$ is the cost.

For the normal distribution $\mathbb{N}(150.5, 49.8)$, use the standard normal CDF for calculation.

Please make an inventory ordering decision: How many wodgets will you order? Please explain your thinking process and decision in detail.
\end{sloppypar}
\end{tcolorbox}

\vspace{3mm}
\noindent {\bf Experiment~3 (High Margin, Uniform Distribution, Risk-Neutral Setting)}
\vspace{3mm}

\begin{tcolorbox}[breakable, colback=gray!5!white, colframe=gray!50!black, title=Risk-Neutral Environment Prompt, width=\textwidth, before skip=10pt, after skip=10pt]
\begin{sloppypar}
You are participating in an inventory decision experiment. You need to make inventory ordering decisions for selling ``wodgets'' products.\\

Experiment rules:\\
- Each wodget sells for 12 francs\\
- Each wodget costs 3 francs\\
- Remaining unsold wodgets can be disposed of for 0 francs\\
- demand is uniformly distributed between 901 and 1200\\
- Your goal is to maximize profit\\

Here is some information that might be helpful:\\
- If sales exceed your order quantity, you will lose some sales opportunities\\
- If the order quantity exceeds sales, you will need to dispose of the remaining inventory\\
- Please consider the costs of ordering too much and ordering too little\\

Note: You have previously learned about the newsvendor problem. In this problem, the profit-maximizing order quantity can be determined by the following formula:
\[
F(q^\star) = \frac{p - c}{p},
\]
where $F(q)$ is the cumulative distribution function, $p$ is the selling price.

For uniform distribution $\mathbb{U}[901, 1200]$, $F(q) = (q - 901)/(1200 - 901)$.

Please make an inventory ordering decision: How many wodgets will you order? Please explain your thinking process and decision in detail.
\end{sloppypar}
\end{tcolorbox}

\end{document}